\documentclass[10pt,twocolumn,letterpaper]{article}

\usepackage[pagenumbers]{cvpr} 

\usepackage{graphicx}
\usepackage{amsmath}
\usepackage{amssymb}
\usepackage{mathrsfs}
\usepackage{mathtools}
\usepackage{booktabs}
\usepackage{rotating}
\usepackage{multirow}
\usepackage{booktabs}
\usepackage{lipsum}
\usepackage[dvipsnames]{xcolor}
\usepackage{enumitem}
\usepackage{inconsolata}
\usepackage{pifont}
\usepackage{soul}
\usepackage{siunitx}
\usepackage{csquotes}
\usepackage{longtable}
\usepackage[accsupp]{axessibility}  

\renewcommand{\paragraph}[1]{\vspace{1mm}\noindent\textbf{#1}}

\newcommand{\bb}{\mathbf{b}}
\newcommand{\bh}{\mathbf{h}}

\newcommand{\bw}{\mathbf{w}}

\newcommand{\bbF}{\mathbf{F}}
\newcommand{\bB}{\mathbf{B}}
\newcommand{\bW}{\mathbf{W}}
\newcommand{\bM}{\mathbf{M}}
\newcommand{\bE}{\mathbf{E}}

\newcommand{\CLS}{\mathsf{CLS}}
\newcommand{\bbR}{\mathbb{R}}

\newcommand{\MN}{\mathcal{N}}\newcommand{\MV}{\mathcal{V}}

\newcommand{\mcT}{\mathcal{T}}

\newcommand{\MC}{\mathcal{C}}

\newcommand{\MP}{\mathcal{P}}
\newcommand{\MB}{\mathcal{B}}
\newcommand{\ML}{\mathcal{L}}
\newcommand{\MG}{\mathcal{G}}

\newcommand{\someone}{\emph{someone}}

\newcommand{\ispice}{iSPICE}
\newcommand{\modelname}{MICap}
\newcommand{\makeblank}{\makebox[1cm]{\hrulefill}}

\definecolor{cvprblue}{rgb}{0.21,0.49,0.74}
\usepackage[pagebackref,breaklinks,colorlinks,citecolor=cvprblue]{hyperref}

\def\paperID{9905} 
\def\confName{CVPR}
\def\confYear{2024}

\title{\modelname: A Unified Model for Identity-aware Movie Descriptions
}

\author{
Haran Raajesh$^{*1}$ \hspace{0.4cm}
Naveen Reddy Desanur$^{*1}$\hspace{0.4cm}
Zeeshan Khan$^2$ \hspace{0.4cm}
Makarand Tapaswi$^1$ \\
$^1$CVIT, IIIT Hyderabad, India \\
$^2$Inria Paris and Département d'informatique de l'ENS, CNRS, PSL Research University \\
{\small
\url{https://katha-ai.github.io/projects/micap/}} \\
{\small
$^*$ denotes equal contribution}
}
\begin{document}

\twocolumn[{%
\renewcommand\twocolumn[1][]{#1}%
\maketitle

\centering
\vspace{-2mm}
\includegraphics[width=\linewidth]{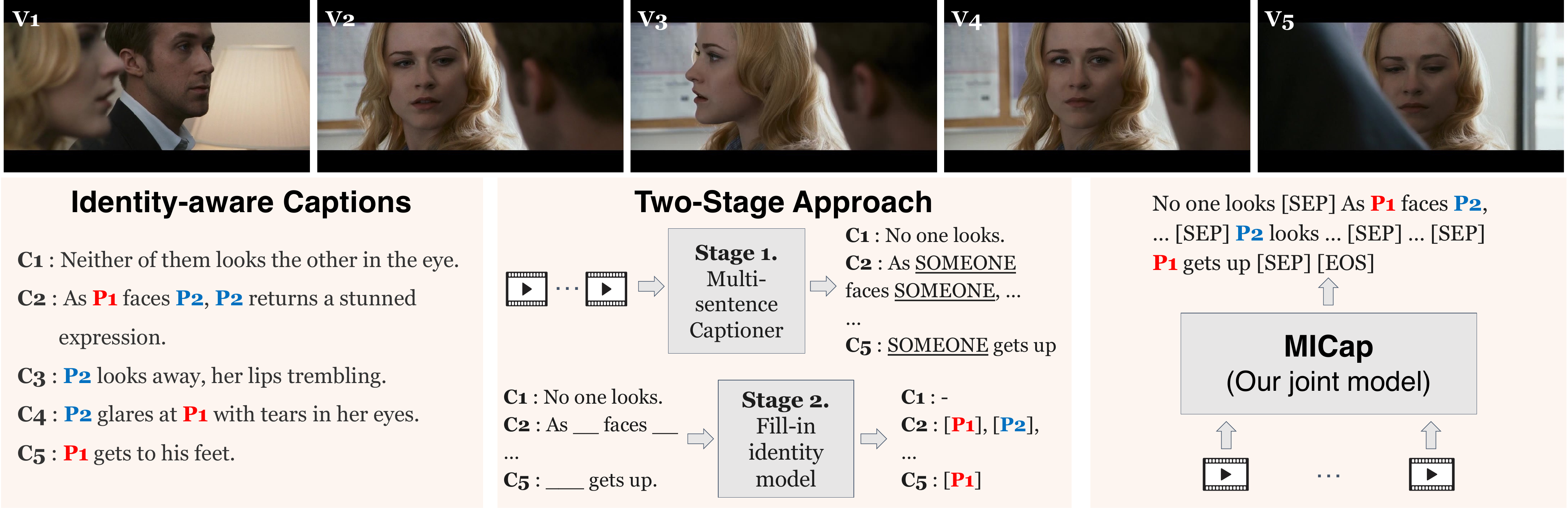}
\vspace{-5mm}
\captionof{figure}{
Identity-aware captioning.
\textbf{Left:}~To understand the story in a set of videos, captions refer to characters by a unique local identifier (\eg~P1, P2, $\ldots$).
The \emph{Fill-in-the-blanks} (FITB) task provides these captions with blanks (removing names) and asks a model to fill local person ids.
\textbf{Middle:}~End-to-end captioning for a videoset is achieved in two stages~\cite{fillin}.
First, captions are generated with \someone{} tags, and then the FITB module is applied to fill-in names.
\textbf{Right:}~We propose a single-stage encoder-decoder id-aware captioning approach that can switch between generating the caption with ids or filling in the ids in a caption, jointly learning from both tasks.}
\label{fig:teaser}
\vspace{6mm}
}]
\maketitle

\begin{abstract}
\vspace{-2mm}
Characters are an important aspect of any storyline and identifying and including them in descriptions is necessary for story understanding.
While previous work has largely ignored identity and generated captions with \someone{} (anonymized names), recent work formulates id-aware captioning as a fill-in-the-blanks (FITB) task, where,
given a caption with blanks, the goal is to predict person id labels.
However, to predict captions with ids, a two-stage approach is required: first predict captions with \someone, then fill in identities.
In this work, we present a new single stage approach that can seamlessly switch between id-aware caption generation or FITB when given a caption with blanks.
Our model, Movie-Identity Captioner (\modelname), uses a shared auto-regressive decoder that benefits from training with FITB and full-caption generation objectives, while the encoder can benefit from or disregard captions with blanks as input.
Another challenge with id-aware captioning is the lack of a metric to capture subtle differences between person ids.
To this end, we introduce \ispice{}, a caption evaluation metric that focuses on identity tuples created through intermediate scene graphs.
We evaluate \modelname{} on Large-Scale Movie Description Challenge (LSMDC), where we show a 4.2\% improvement in FITB accuracy, and a 1-2\% bump in classic captioning metrics.
\end{abstract}

\vspace{-2mm}
\section{Introduction}
\label{sec:intro}

Building computer vision models that understand the story of a movie is a long-standing challenge.
A step towards this is movie description~\cite{lsmdc,mvad,mpiimd}.
Given a short clip of 2-5 seconds, models are required to generate a caption that describes the visual scene.
Captions in the Large Scale Movie Description Challenge (LSMDC)~\cite{lsmdc}, a combination of~\cite{mpiimd,mvad}, are obtained from \emph{audio descriptions} (AD) that are used to convey the (visual) story to a visually impaired audience.
The original version of the LSMDC challenge suggests captioning a single clip and anonymizes all character names with \someone.

While using the \someone{} tag to describe a character's activity in a single video is acceptable, the lack of identity continuity across a \emph{videoset} (group of $N$ consecutive videos) hampers understanding.
To remedy this,
Pini~\etal~\cite{mvadnames} extend MVAD~\cite{mvad} as \emph{MVAD names} where character names are predicted by linking to the appropriate face detection/track; and
Park~\etal~\cite{fillin} propose a fill-in-the-blanks (FITB) task to replace \someone{} tags with local cluster identities (\eg~P1, P2, $\ldots$) in a videoset (\cref{fig:teaser} left).

The latter approach~\cite{fillin} provides two advantages:
(i)~it does not require time-consuming ground-truth annotations linking faces and blanks~\cite{mvadnames}; and
(ii)~using local cluster ids helps convey the story%
\footnote{Note, cluster ids can be easily mapped to gender- and culture-appropriate names instead of using P1, P2, $\ldots$ for storytelling.}
without the need for models with world knowledge (CLIP~\cite{clip}, GPT~\cite{gpt2}, \etc) or an IMDb castlist with photographs~\cite{autoad2}, making the approach applicable to indie films or home-edited videos.

To generate id-aware captions, \cite{fillin} proposes a two-stage approach shown in \cref{fig:teaser} (middle).
The first stage~\cite{adversarial} ingests a videoset and generates a \emph{captionset} (a set of $N$ captions, one for each video) using the \someone{} tags;
while the second stage replaces \someone{} with appropriate local person id labels.
While the two-stage setting unites the two worlds of video description and character identification, it is not ideal as errors in captioning may adversely affect FITB as both methods are modeled independently.
In this work, we propose a single-stage approach (\cref{fig:teaser} right)
that can seamlessly switch between both tasks.

\paragraph{Challenges with Fill-In.}
For the FITB task,
\cite{fillin} encodes blanks in the ground-truth (GT) captionset using bidirectional context through the BERT encoder.
These blanks attend to the face features clustered within a single video, not accounting for other faces coming from the videoset.

Using the blank representations, the person ids are predicted in an auto-regressive manner.

We note some disadvantages with this approach:
(i)~Faces are clustered within each video. This means identity information across videos is not directly observed by the model.
(ii)~When a character is mentioned in the caption, their face need not be present in the clip (\eg~\cref{fig:teaser} left, C4 and C5 mention P1 whose face is turned and not visible).
(iii)~BERT-based blank embeddings provided at the encoder are unable to capture face information properly, resulting in a model that largely focuses on text embeddings to solve FITB (\eg, in~\cite{fillin}, FITB accuracy only improves by 1.5\% (64.4 to 65.9) with visual inputs).

\paragraph{Proposed model benefits.}
We overcome these problems using a new paradigm for id-aware multi-video description through a single-step sequence-to-sequence model.
We unify the two tasks of FITB and caption generation, by auto-regressively unrolling the descriptions along with their local character ids, via a Transformer based encoder-decoder model.
Our model, dubbed as the \emph{Movie-Identity Captioner} (\modelname), enables joint training and independent evaluation for both tasks:
(i)~given only the videoset, our model generates an id-aware captionset; and
(ii)~when a captionset with \someone{} tags exists, our model fills in local identities.

To overcome text-only shortcuts, we propose auto-regressive decoding of the full caption even for FITB and show that our multimodal model outperforms a text-only model significantly.
We teacher force the ground-truth caption containing the blanks (person ids), and predict one token at a time using causal masking.
Note, learning happens only at select tokens where person id labels are predicted.
This way the model (decoder) learns to sequentially use the GT (teacher forced) caption for the FITB task with uni-directional (causal) attention.
During inference, we switch between the two tasks by deciding whether the decoder is teacher forced with a given captionset or not.

\paragraph{Identity-aware evaluation.}
Existing captioning metrics like CIDEr~\cite{cider} and BLEU~\cite{bleu} do not account for identity sensitive descriptions.
For example \emph{``P1 is walking towards P2"} and \emph{``P2 is walking towards P1"} will result in high n-gram based scores due to common middle words.
We propose a new identity-aware caption evaluation metric \emph{\ispice}.
Specifically, we are motivated by SPICE's~\cite{spice} ability to parse a caption into a scene graph, and match a predicted caption with ground-truth based on similarity across generated tuples.
To compute \ispice{}, we intervene in this process and remove tuples not associated with a person label before computing the F1 scores.

\paragraph{Contributions.}
In summary,
(i)~we propose a new paradigm for identity-aware multi-sentence movie description using a single-stage approach that unifies FITB with full caption generation.
(ii)~We formulate this task as an auto-regressive sequence-to-sequence generation that is able to describe the video and use local person id labels across a videoset (multiple videos).
We show that joint training improves knowledge sharing and boosts performance.
(iii)~We enable seamless task switching allowing independent evaluation of
(a)~caption generation with identities, and (b)~filling in identity labels given a caption.
(iv)~We propose a new identity-aware captioning metric, \ispice{}, that extends SPICE, and show its sensitivity to identities while evaluating captions.
(v)~Finally, \modelname{} improves over the state-of-the-art for FITB by 4.2\% and identity-aware captioning by 1.4\% CIDEr and 1.8\% METEOR.

\section{Related Work}
\label{sec:relwork}

We address related work from three areas:
(i)~video captioning at large,
(ii)~identity-aware captioning, and
(iii)~metrics used for evaluating captions.

\paragraph{Video captioning}
has gained a lot of attention since the advent of deep learning.
The typical task is to generate a single sentence description for a trimmed video, and is formulated as a sequence-to-sequence problem~\cite{lrcn, s2svt, tvnl, eterqa, swinbert, univl, seoend}.
A more challenging setup is multi-sentence generation, typically applied to longer videos and requires long-term temporal consistency~\cite{adversarial, rohrbachcoherent, shin2016beyond, yu2016video}.
Video situation recognition, VidSitu~\cite{vidsitu, GVSR} presents a structured alternative where multiple captions are generated per event based on the semantic role labeling framework.

Different from multi-sentence captioning, \emph{dense video captioning}, requires temporally localizing and generating captions for every event in an untrimmed video~\cite{krishna2017dense, wang2021end, zhou2018end, vid2seq}.
While most approaches for dense video captioning use a 2-stage approach, \ie~temporal localization with event proposals then event captioning~\cite{ krishna2017dense, wang2018bidirectional, wang2020event}, recent methods, jointly model the two tasks for better temporal consistency~\cite{ Chadha2020iPerceive, chen2021towards,  deng2021sketch, li2018jointly, mun2019streamlined, rahman2019watch, shen2017weakly, shi2019dense, wang2018bidirectional, wang2021end, zhou2018end}.
The state-of-the-art, PDVC~\cite{wang2021end}, learns DETR-style event queries and performs localization and captioning over each query using 2 separate heads.
Recently, Vid2Seq~\cite{vid2seq} proposed to further unify the two tasks by using a single sequence-to-sequence model and generating both the localization and captions with a single auto-regressive Transformer decoder.
Similar to above ideas, we unify two seemingly different tasks of character identification and description by formulating them as an auto-regressive sequence generation task.

\paragraph{Id-aware captioning datasets.}
None of the above works focus on person identity while generating captions.
VidSitu~\cite{vidsitu}, perhaps the closest, contains references to people by descriptions such as \emph{man in a black jacket}.
This is an issue when the domain is movie description~\cite{lsmdc,mvad}, where identities are anonymized to \someone{} which hinders building
practical applications like \emph{Audio Descriptions}~\cite{autoad} for visually impaired users.
While~\cite{mvadnames} links character names in descriptions with face tracks, they require significant annotation effort that is not scalable.
A more recent Movie Audio Description dataset, MAD~\cite{MAD}, is a popular source for movie descriptions.
But it uses real names that require models with world knowledge. 
Different from above, Park~\etal~\cite{fillin} propose identity-aware captioning as a fill-in-the-blanks task where they assign local person ids (cluster ids) to characters appearing in 5 consecutive video clips.
We adopt this setting for our work.

\paragraph{Id-aware captioning methods.}
Identity-aware captioning is a challenging task that has recently started to attract attention.
Among the first works, \cite{fillin} proposes a 2-stage pipeline of first captioning with identities anonymized as \someone{} using a multi-sentence captioning model~\cite{adversarial}, followed by learning an identity prediction FITB model that fills in the \someone{} with local person identities.
However, as discussed in the introduction (Challenges with Fill-In), the specific 2-stage approach suffers from several disadvantages.
Different from~\cite{fillin}, we propose a single stage sequence-to-sequence model, that outperforms the 2-stage approach.
In this area, another work~\cite{cisin} requires ground-truth mapping between person identities (blanks) in the description to face tracks in the videos.
However, this approach is not scalable.
Very recently, AutoAD-II~\cite{autoad2} proposed to generate movie descriptions with proper names, on the MAD~\cite{MAD} dataset.
While innovative, this approach requires additional IMDb castlist information with photographs.
While modeling proper names directly is useful, tagging names to unique person ids in a local videoset is possible and is the motivation for works on person  clustering~\cite{brown2021cluster,ballclustering} as opposed to person identification~\cite{knockknock,sherlock_personid}.

\paragraph{Caption evaluation metrics}
are typically based on n-gram matching, with few differences.
CIDEr~\cite{cider}, BLEU~\cite{bleu}, and METEOR~\cite{meteor} all evaluate n-gram similarities between a single or multiple candidate references and the generated caption.
Recently, Large Language Models (LLMs) are used for reference-based (\eg~BERTScore~\cite{bertscore}, CLAIR~\cite{clair}) or
or Large Vision-Language Models (VLMs) for reference-free caption evaluation (\eg~CLIP Score~\cite{clipscore}).
However, model-based metrics may be difficult to interpret, and also require the model to be sensitive to identities.
Different from both directions, SPICE~\cite{spice} evaluates captions by first transforming them into a scene graph and analyzing presence of shared tuples between the predicted and ground-truth (reference) captions.
However, none of the metrics reliably evaluate identity-aware captions, as a robust metric should be sensitive to identity manipulations (swap/add/remove). 
We propose a new metric \ispice{} that focuses primarily on person-identity specific semantics.

\section{Method}
\label{sec:method}

\begin{figure*}[t]
\centering
\includegraphics[width=\linewidth]{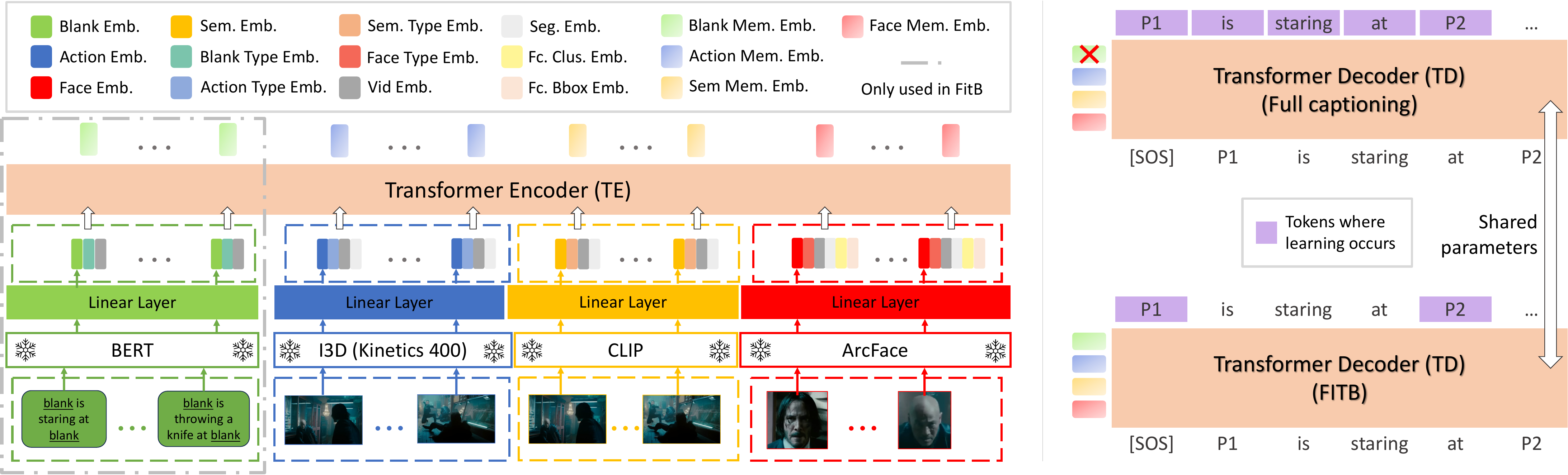}
\vspace{-3mm}
\caption{Identity-aware captioning.
\textbf{Left}:~illustrates the Transformer Encoder used to capture multimodal inputs such as text (blanks), action, semantic, and face. These tokens are used as memory for the Transformer Decoders.
\textbf{Right}:~the same Transformer Decoder can be used for both tasks of full caption generation and fill-in-the-blanks (FITB).
The model is trained end-to-end with losses applied to tokens indicated in purple.
Text tokens are not presented to the decoder for full caption generation.
Joint training improves knowledge sharing resulting in performance improvements.
}
\vspace{-2mm}
\label{fig:model}
\end{figure*}

We present a single-stage sequence-to-sequence approach for identity-aware fill-in-the-blanks (FITB).
Later, we will show that this architecture can be easily re-purposed for generating video descriptions.

\paragraph{Notation.}
Before we start, we define some notation.
For the rest of this section, we will operate with a videoset $\MN$ consisting of $N$ video clips $V_i$ and corresponding captionset $\MC = \{C_i\}_{i=1}^N$, where $C_i$ describes video $V_i$.
As both sets come from consecutive videos, it is very likely that same characters appear across them.
As an example, consider the videoset frames and captionset shown in \cref{fig:teaser}.

\subsection{Auto-regressive FITB}

In FITB, we replace each person-id (P1, P2, $\ldots$) with a blank.
We denote $\hat{\MC}$ as the captionset with $\MB$ blanks.
Formally, we define the captionset as a sequence of $L$ words $[w_j]_{j=1}^L$, some of which have been converted to blanks $\{b_k\}_{k=1}^{|\MB|}$.
The goal of our model is to fill each blank with the correct person-id label from the set $\MP = \{P_l\}_{l=1}^{|\MP|}$.
Note, the person-id labels are reusable across videosets, \ie~a character only needs to be referred consistently by the same identity within a videoset.

We present Movie-Identity Captioner (\modelname), an auto-regressive Transformer encoder-decoder model for filling person blanks.
\modelname{} consists of two parts:
(i)~Feature extractors and a Transformer encoder to build the captioning memory (\cref{fig:model} left); and
(ii)~A Transformer decoder that switches between FITB or full captionset generation
(\cref{fig:model} right).
For clarity, we will highlight differences to prior work~\cite{fillin} throughout this section.

\subsubsection{Creating the Captioning Memory}

\paragraph{Visual feature extraction.}
We extract 3 features from the videoset to capture semantic, action, and face information.

\underline{Semantic embeddings}
are captured using CLIP~\cite{clip}.
From each video $V_i$, we sub-sample frames $f_{it}$ at 5 fps and encode them with the CLIP image encoder.
For efficient batching, we truncate or pad to $T{=}50$ frames per video, and stack them to create semantic features $\bbF^\text{s} \in \bbR^{NT \times d^\text{s}}$.

\underline{Action embeddings}
are captured using I3D~\cite{i3d}.
Similar to~\cite{fillin}, each video is divided into $S{=}5$ segments, and features within each segment are mean pooled.
We stack features across the videoset to obtain $\bbF^\text{a} \in \bbR^{NS \times d^\text{a}}$.

\underline{Faces}
are detected using Retina Face~\cite{retina} and represented using Arcface~\cite{arcface}.
Across the videoset, we collect a maximum of $F{=}300$ face detections.
With each face detection, we associate the video index $i$ (for $V_i$) from which it is derived and a normalized spatial bounding box location.
We stack features to obtain $\bbF^\text{f} \in \bbR^{F \times d^\text{f}}$.

We bring all these features to a common $d$ dimensional space using separate linear projection layers for each modality:
$\bW^\text{mod} \in \bbR^{d \times d^\text{mod}}$,
where mod takes on values:
\underline{s} for semantic,
\underline{a} for action, and
\underline{f} for face.

\paragraph{Captionset feature extraction.}
Similar to~\cite{fillin}, we also extract blank embeddings by feeding the captionset to BERT (fine-tuned for gender prediction as in~\cite{fillin}) and using the contextualized tokens:
\begin{equation}
\small
[\hat{\CLS}, \hat{\bw}_1, \ldots, \hat{\bb}_k, \ldots] = \text{BERT}([\CLS, w_1, \ldots, b_k, \ldots]) \, .
\end{equation}
The blank embedding is a concatenation of contextualized tokens:
$\bb_k = [\hat{\CLS}, \hat{\bb}_k]$.
We stack these to create a matrix
$\bB \in \bbR^{|\MB| \times 2\cdot d^\text{bert}}$
and transform them to the same space through a linear projection $\bW^\text{bert} \in \bbR^{d \times 2\cdot d^\text{bert}}$.

\paragraph{Face clustering.}
Instead of creating face clusters within each video and using blank embeddings to attend to them (as done in~\cite{fillin})
we adopt a soft approach for incorporating cluster information in \modelname.
First, we perform clustering using DBSCAN across \emph{all} $F$ detections in the \emph{videoset}, resulting in $\MG$, a set of face groups.
This allows our model to associate faces across videos as the same or different person.
Next, we prevent propagating errors caused by clustering and mean pooling representations by adding a cluster-id based learnable embedding $\bE^\text{fcl}$ to the face representations.

\paragraph{Additional embeddings}
are added to various features to orient the model:
(i)~$\bE^\text{typ} \in \bbR^{d \times 4}$ disambiguates between the 4 types of features.
(ii)~$\bE^\text{vid} \in \bbR^{d \times N}$ consists of $N$ embeddings to inform the model of the source video index for any visual or blank token.
(iii)~$\bE^\text{seg} \in \bbR^{d \times S}$, together with $\bE^\text{vid}$, allows to localize any feature to the correct video and segment.
(iv)~$\bE^\text{fcl} \in \bbR^{d \times |\MG|}$ is the face cluster index embedding described above, and
(v)~$\bE^\text{bbox} \in \bbR^{d \times 4}$ transforms normalized face detection bounding box coordinates to provide the model spatial information.

We create input tokens as follows (with appropriate indexing hidden for brevity):
\begin{align}
\hat{\bB} &= \bW^\text{bert} \bB + \bE^\text{typ}_0 + \bE^\text{vid} \, , \\
\hat{\bbF}^\text{s} &= \bW^\text{s} \bbF^\text{s} + \bE^\text{typ}_1 + \bE^\text{vid} + \bE^\text{seg} \, , \\
\hat{\bbF}^\text{a} &= \bW^\text{a} \bbF^\text{a} + \bE^\text{typ}_2 + \bE^\text{vid} + \bE^\text{seg} \, , \\
\label{eq:face_clust_embed}
\hat{\bbF}^\text{f} &= \bW^\text{f}\bbF^\text{f} + \bE^\text{typ}_3 + \bE^\text{vid} + \bE^\text{seg} + \bE^\text{fcl} + \bE^\text{bbox} .
\end{align}

\paragraph{A Transformer encoder (TE)}~\cite{transformer} of $L_E$ layers is used to combine and refine individual representations mentioned above.
Thus, the final memory bank is:
\begin{equation}
\bM = [\tilde{\bB}, \tilde{\bbF}^s, \tilde{\bbF}^a, \tilde{\bbF}^f]
= \text{TE}(
[\hat{\bB}, \hat{\bbF}^s, \hat{\bbF}^a, \hat{\bbF}^f]
) \, .
\end{equation}

\subsubsection{Auto-regressive Identity Prediction}
We now present the process of filling blanks.
Similar to the encoder, we use a couple embeddings for the decoder.
(i)~$\bE^\text{vid}$ (shared with encoder) informs the decoder of the video index that is being captioned; and
(ii)~$\bE^\text{pos}$ encodes learnable position embeddings similar to the original Transformer~\cite{transformer}.
We use the memory embeddings extracted from the video as key-value pairs and blanks in the Transformer decoder (TD) as queries.
Given a captionset $\hat{\MC}$, we generate the next word as
\begin{align}
\bh_{j+1} &= \text{TD}([w_1, \ldots, w_j]; \bM) \, , \\
w_{j+1} &= \arg\max_\MV \bW^{\MV} \bh_{j+1} \, .
\end{align}
$\bh_{j+1}$ represents the output of TD at the $j+1^\text{th}$ timestep and is obtained through a series of $L_D$ decoder layers that compute
self-attention to previous words,
and cross-attention to the memory.
$\bW^\MV$ is a linear classifier in $\bbR^{\MV \times d}$, where $\MV$ is the word vocabulary.

For the FITB task, the captionset already contains the correct caption words.
Thus, the output prediction is relevant only when $w_{j+1}$ is a blank $b_k$.
In such a case, we can use a smaller output classifier $\bW^\MP$ that picks one among $\MP$ person-id labels.
We rewrite the above equations as:
\begin{align}
\label{eq:id_hidden}
\bh_{j+1} &= \text{TD}([w_1, \ldots, w_j]; \bM) \, , \\
\label{eq:id_pred}
w_{j+1}{=}\hat{y}_k &= \arg\max_\MP \bW^{\MP} \bh_{j+1} \, ,
\end{align}
where $\hat{y}_k \in \MP$ is the predicted person-id label for blank $b_k$.

\paragraph{Training and inference.}
We train \modelname{} by applying a cross-entropy loss at every blank:
\begin{equation}
\label{eq:loss_fitb}
\ML_\text{FITB} = - \sum_{k=1}^{|\MB|} y_k \log \text{softmax}_\MP \left( \bW^\MP \bh_{j+1} \right) \, ,
\end{equation}
where $y_k$ is the correct label for blank $b_k$.
The key difference to~\cite{fillin} is that our decoder observes each word of the captionset in an auto-regressive manner.

During inference, we simply follow \cref{eq:id_pred} to compute person-id label predictions for blanks in a captionset.

\subsection{Joint Fill-in and Captioning}

We first present how \modelname{} can be adapted for generating the entire captionset.
Then, we will present the opportunity of joint training.

\paragraph{From FITB to generating the captionset.}
In this scenario, the model is shown the videoset $\MN$ and expected to generate an id-aware captionset $\MC$.
We make two small changes:

(i)~The memory bank is restricted to visual features,
$\bM = [\tilde{\bbF}^s, \tilde{\bbF}^a, \tilde{\bbF}^f]$.
In fact, we cannot compute blank embeddings $\tilde{\bB}$ as the captionset needs to be predicted.

(ii)~When decoding the next word of the captionset, we use an augmented vocabulary consisting of normal language tokens (from $\MV$) and person-id labels (from $\MP$).
We predict the next word as shown below:
\begin{align}
\label{eq:aug_vocab}
\MV^* &= \MV + \MP \, , \\
\bh_{j+1} &= \text{TD}([w_1, \ldots, w_j]; \bM) \, , \\
\label{eq:pred_everyword}
\hat{w}_{j+1} &= \arg\max_{\MV^*} \bW^{\MV^*} \bh_{j+1} \, ,
\end{align}
and train our model to minimize
\begin{equation}
\label{eq:loss_cap}
\ML_{cap} = - \sum_{j=1}^L w_{j+1} \log \text{softmax}_{\MV^*} \left( \bW^{\MV^*} \bh_{j+1} \right) \, .
\end{equation}
We can use \cref{eq:pred_everyword} during inference to predict the entire captionset until the end-of-sentence token is triggered.

\paragraph{Joint training.}
Can we train the same instance of \modelname{} to generate the captionset and fill-in-the-blanks with identity information?
Yes, we suggest an efficient way to do so.

Given a batch of data consisting of multiple paired videosets and captionsets $(\MN, \MC)$, we forward it through the model twice.
In the first forward pass, we replace the person-id labels with blanks, \ie~create $\hat{\MC}$, and compute losses and gradients to predict the blank's labels (see \cref{eq:loss_fitb}).
In the second forward pass conducted on the same batch, we assume that $\MC$ is not available as input and use the augmented vocabulary $\MV^*$ to compute loss and gradients for each word as in \cref{eq:loss_cap}.
We can either accumulate gradients and optimize parameters at the end of both forward passes or optimize parameters after each pass.

Note, the classifier parameters $\bW^\MP$ are subsumed under $\bW^{\MV^*}$.
We find that sharing the classifier $\bW^{\MV^*}$ for both forward passes works best.

Thus, we unite seemingly disparate tasks of filling in person-id labels in blanks and generating the full captionset in a single model with a single set of parameters.

\section{Identity-aware SPICE}
\label{sec:metric}

Inspired by a metric used in image captioning evaluation called Semantic Propositional Image Caption Evaluation (SPICE)~\cite{spice}, we propose a new metric --
identity-aware SPICE (\ispice{} for short) --
to evaluate the quality of video descriptions, especially pertaining to identity labels.

\paragraph{Why SPICE?}
The classic captioning metrics borrowed from language translation such as BLEU~\cite{bleu}, ROUGE~\cite{rouge}, METEOR~\cite{meteor}, and CIDEr~\cite{cider} rely primarily on n-gram overlap.
However, as indicated in~\cite{spice},
``n-gram overlap is neither necessary
nor sufficient for two sentences to convey the same meaning''.
SPICE is shown to have a high correlation with human judgement (0.88) as compared to METEOR (0.53) or CIDEr (0.43) on the MS-COCO image captioning dataset~\cite{spice}.

\paragraph{How is SPICE calculated?}
SPICE estimates quality of a caption in two stages.
First, the reference and predicted caption are converted to \emph{scene graphs}~\cite{scenegraph, imagescene} that explicitly encode objects, attributes, and relationships.
This abstraction provides a list of tuples $\mcT_r$ and $\mcT_p$ for the reference and predicted captions.
SPICE is the F1-score that measures logical conjunction (overlap):
\begin{equation}
\text{SPICE} = \text{F}_1(\mcT_r, \mcT_p) \, .
\end{equation}

\paragraph{\ispice{}}
is a simple modification of SPICE.
We intervene at the list of tuples and filter out tuples that do not have at least one character identity.
We define
\begin{equation}
\text{\ispice} = \text{F}_1(\mcT_r^{p2+}, \mcT_p^{p2+}) \cdot
\text{F}_1(\mcT_r^{p1}, \mcT_p^{p1}) \, ,
\end{equation}
where $\mcT_r^{p2+}$ denotes the list of tuples with a person-id label having 2 or more elements and $\mcT_r^{p1}$ is a set of person-id labels in the reference captionset.
The first term scores whether the correct person-id label is used together with a verb or attribute, while the second term checks that the total number of person-id labels match..
A couple examples of the matching process are presented in the supplement.

\paragraph{Validation.}
We validate \ispice{} by an experiment that measures sensitivity to changes in identity.
Given a reference captionset, we compare it against itself to obtain a base score $s$.
Next, we modify the reference captionset by swapping, adding new, or removing existing id labels.

\textbf{1. Swapping:}
Here, id tokens are replaced with another id present in the captionset.
The number of these tokens is selected at random for each captionset.
We first identify \textit{eligible id} tokens whose ids are present more than once in the captionset.
This is done to prevent the case where standalone ids are selected and replaced with each other that does not change the meaning.
For example, the caption \textit{P1 carries P2} is equivalent to \textit{P2 carries P1} if P1 and P2 are not re-used elsewhere in the captionset.
When the id occurs multiple times, \eg~\textit{P1 carries P2. P2 is unconscious}, the replacement \textit{P2 carries P1. P2 is unconscious} changes the meaning of the story.
Once these eligible tokens are identified, a random subset is replaced with another id present in the captionset to generate the modified caption.

\textbf{2. Addition:}
Here, we select an id token at random and change it to an id token that is not present in the current captionset, adding new identities.
Again, we do not replace tokens whose id appears only once.

\textbf{3. Removal:}
Here, we replace a single occurrence id token (chosen at random) with an id token that exists in the captionset, thereby removing the identity.

\begin{table}[t]
\small
\centering
\tabcolsep=0.12cm
\begin{tabular}{l ccccccc}
\toprule
Experiments   & iS        & S & B4 & C & M & R & BSc \\
\midrule
Swapping      & \textbf{0.55} & 0.85  & 0.87  & 0.86  & 0.61   & 0.95  & 0.99 \\
Addition & \textbf{0.51} & 0.86  & 0.89  & 0.88  & 0.6    & 0.95  & 0.99 \\
Removal      & \textbf{0.46} & 0.84  & 0.87  & 0.86  & 0.6    & 0.95  & 0.99 \\
\bottomrule
\end{tabular}
\vspace{-2mm}
\caption{Sensitivity of metrics to id manipulation in the original caption. \ispice{} shows highest reduction in performance when replacing, adding, or removing ids, indicating that it is a good metric for id-aware captioning
iS=\ispice, S=SPICE, B4=BLEU4, C=CIDEr, M=METEOR, R=ROUGE, BSc=BERTScore.
}
\label{tab:new_metric}
\vspace{-5mm}
\end{table}

\paragraph{Id normalization.}
Prior to scoring, a normalization operation is performed on the captionset.
The first unique id label is set to P1, the second to P2 an so on.
This ensures that the captionsets \textit{P2 carries P1} or \textit{P4 carries P3}, are treated as the same captionset \textit{P1 carries P2}.

\paragraph{Results.}
We compute a new score $\hat{s}$ for each edited captionset by comparing it against the reference.
We report the drop in performance $\hat{s} / s$ as the sensitivity of a metric to changing identities.
We create 3 manipulated samples for each type and report averaged scores over all 1443 captionsets from the validation set in~\cref{tab:new_metric}.
We observe that \ispice{} obtains the smallest score, indicating the highest sensitivity to manipulating identities, a desirable property.

\section{Experiments}
\label{sec:experiments}

We present experiments on the LSMDC~\cite{lsmdc} dataset in the identity-aware multi-video captioning setup~\cite{fillin}.
We describe the experimental setup first, followed by implementation details and metrics.
The evaluation is presented for
(i)~Fill-in-the-blanks and
(ii)~Identity-aware captioning.

\subsection{Setup}

\paragraph{Dataset.}
LSMDC consists of 128,118 short video clips extracted from 202 movies.
Each video has a caption, either from the movie script or from transcribed DVS (descriptive video services) for the visually impaired.
The median video duration is \SI{3}{\second}, average is \SI{4.2}{\second}, and std dev is \SI{3.1}{\second}.
The dataset is split into 101,079 clips for training, 7,408 for validation, 10,053 for public test, and 9,578 for blind test.
We report and compare results on the validation set as the test set labels are not released and the evaluation server is down.

In the Fill-in challenges, the movie descriptions are evaluated on sets of 5 clips taken at a time.
Characters are identified across the clips to provide meaningful narratives.
The training videosets use overlapping clips (\eg~1-5, 2-6) for data augmentation but the val and test videosets are non-overlapping.
We train on 98,527 videosets and report results on 1,443 val videosets.
All three tasks of the LSMDC challenge~\cite{lsmdc} are evaluated on the same sets of 5 clips.
We focus on task 2: filling in local person ids; and
task 3: description generation with local character IDs.

\paragraph{Implementation details.}
Videosets have $N{=}5$ clips, we set the captionset length to 120 tokens.
The hidden dimension for encoder and decoder in \modelname{} is $d{=}512$, and we use $L_E{=}2$ and $L_D{=}3$ layers.
We train our model with a learning rate of $5{\times} 10^{-5}$ for 30 epochs.
The vocabulary sizes are $|\MP|{=}11$ and $|\MV|{=}30522$.
We train on one RTX 2080 GPU with a batch size of 16 videosets/captionsets.

\begin{figure*}[t]
\includegraphics[width=\linewidth]{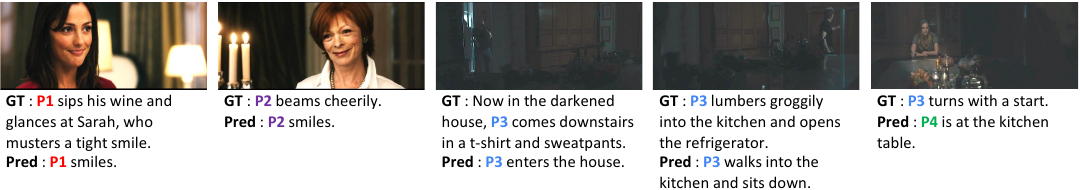}
\vspace{-6mm}
\caption{We show a qualitative example of our joint training approach.
The dataset is highly challenging, with shot changes and dark scenes that are typical in movies.
Yet our model is able to perform reasonably well in this example.
While the predicted captions (Pred) are different from the ground-truth (GT), they capture the overall meaning.
\modelname{} predicts diverse ids correctly in this case and does not overfit to only predicting P1, or P1 and P2.
In fact, in the last clip, as P3 turns (indicated in GT), we see P4 sitting at the table (indicated in Pred), which is a correct caption!
The last clip also highlights challenges of evaluating captions correctly.
}
\label{fig:qualitative_eval}
\vspace{-4mm}
\end{figure*}

\paragraph{Fill-in metrics.}
For the Fill-in task we evaluate results using all pairs of blanks in the captionset as proposed by~\cite{fillin}.
Pairs that require both ids to be same are called are evaluated with same accuracy (``Same-acc'').
Different id pairs are evaluated using ``Diff-acc''.
``Inst-acc'' is the combined accuracy while ``Class-acc'' computes the harmonic mean.

\paragraph{Captioning metrics.}
We use METEOR~\cite{meteor}, CIDEr~\cite{cider}, SPICE~\cite{spice} and our newly proposed metric \ispice{} to evaluate the quality of our generated captions.

\subsection{Evaluating on the Fill-in Task}

\paragraph{\modelname{} makes better use of visual features.}
In \cref{tab:fillin_ablations},
our text-only model (row 2) is comparable to~\cite{fillin}'s text-only (R0).
While~\cite{fillin} improves by 1.5\% (R1), \modelname{} achieves a significant 4.7\% improvement (R6).

\paragraph{Ablations on visual features.}
\cite{fillin} computes face clusters within a video and provides mean pooled features of faces in a cluster.
R3 of \cref{tab:fillin_ablations} uses these features in \modelname{} (with embeddings from \cref{eq:face_clust_embed}).
The only decoder model (only-dec) achieves a 0.6\% improvement, while the encoder-decoder model (enc-dec) shows 1.4\% improvement over R1.
Next, in R4, we swap out face cluster features to individual face detections, while still using FaceNet for a fair comparison; but using embeddings as shown in \cref{eq:face_clust_embed}.
This improves the only-dec model by a further 0.9\%, but enc-dec shows negligible change.
We incorporate CLIP features as additional tokens in the memory, resulting in a 0.35\% increase in enc-dec (R5).
Finally, in R6, swapping FaceNet~\cite{facenet} to Arcface~\cite{arcface} results in a relatively large improvement of 1.6\% (only-dec) and 1.4\% (enc-dec).

\begin{table}[t]
\small
\tabcolsep=0.13cm
\centering
\begin{tabular}{ll cc}
\toprule
\# & Method & Only Dec & Enc-Dec \\
\midrule
0 & FillIn text-only~\cite{fillin} & - & 64.4 \\
1 & FillIn multimodal~\cite{fillin} & - & 65.9  \\
2 & \modelname{} text-only & - & 64.45 \\
3 & \modelname{} w face clusters of~\cite{fillin} & 66.56 & 67.29 \\
4 & \modelname{} w raw face detections & 67.48 & 67.35 \\
5 & \modelname{} 4 + w CLIP features & 67.38  & 67.70  \\
6 & \modelname{} 5 + w Arcface features & \textbf{68.94} & \textbf{69.14} \\
\bottomrule
\end{tabular}
\vspace{-3mm}
\caption{Ablation study showing the impact of various inputs on the decoder only and encoder + decoder model.
We report \textit{class accuracy} as a single metric for comparison.
}
\label{tab:fillin_ablations}
\vspace{-3mm}
\end{table}

\begin{table}[t]
\small
\tabcolsep=0.10cm
\centering
\begin{tabular}{l cccc}
\toprule
Method & Same & Different & Instance & Class \\
\midrule
\multicolumn{5}{c}{\textbf{Test set}} \\
Yu~\etal~\cite{yu19lsmdc} & 26.4 & 87.3 & 65.9 & 40.6  \\
Brown~\etal~\cite{brown19lsmdc} & 33.6 & 81.0 & 64.8 & 47.5  \\
FillIn text-only~\cite{fillin} & 56.0 & 71.2 & 64.8 & 62.7  \\
FillIn~\cite{fillin} & 60.6 & 70.0 & 69.6 & 64.9 \\
\midrule
\multicolumn{5}{c}{\textbf{Validation set}} \\
FillIn~\cite{fillin} & 63.5 & 68.4 & 69.0 & 65.9  \\
Ours (only-dec) & 65.1 & \textbf{73.3} & 73.0 & 68.94  \\
Ours (enc-dec) & \textbf{65.7} & 72.9 & \textbf{73.0} & \textbf{69.14}  \\
\bottomrule
\end{tabular}
\vspace{-3mm}
\caption{Comparison to SotA on fill-in-the-blanks (FITB, task 2) of the LSMDC challenge.}
\label{tab:sota-fillin}
\vspace{-2mm}
\end{table}
\paragraph{SotA comparison.}
\cref{tab:sota-fillin} reports results on all 4 FITB metrics.
As we do not have access to the test set labels and the evaluation server is inactive, we use FillIn's results as a proxy for comparison.
First, in the top half, we see that FillIn~\cite{fillin} outperforms other works.
In the bottom half, on the validation set, we compare our approach against FillIn showing a significant improvement of 4\% on instance accuracy and 3.2\% on class accuracy.
As we teacher force captions through the decoder, our only decoder model also outperforms~\cite{fillin} by 3\% on class accuracy.

\begin{table}[t]
\small
\tabcolsep=0.12cm
\centering
\begin{tabular}{l cccc c}
\toprule
& \multicolumn{4}{c}{Captioning metrics} & FITB \\
Method & C & M & S & iS & Class Acc.\\
\midrule
FITB only & - & - & - & - & 69.14  \\
Full caption only & 8.01 & 12.29 & 13.11 & 0.777 & - \\
\midrule
Joint training & \textbf{9.09} & \textbf{12.47} & \textbf{13.30} & \textbf{0.788} & \textbf{70.01} \\
\bottomrule
\end{tabular}
\vspace{-3mm}
\caption{Ablation showing joint training is better than performing FITB or full captioning separately.
Captioning metrics are C=CIDEr, M=METEOR, iS=\ispice, S=SPICE.}
\label{tab:ablation_joint_training}
\vspace{-4mm}
\end{table}

\subsection{Evaluating Joint Fill-in and Captioning}
We evaluate \modelname{} trained jointly for FITB and id-aware caption generation.
\cref{tab:ablation_joint_training} shows that joint training on fill-in and captioning improves the performance on both the tasks.
Class accuracy on FITB improves by 0.9\% and captioning metric CIDEr by 1\%.
We also see a small 0.01\% improvement in \ispice{}, which we think is important considering the difficulty of the metric.
This suggests that both the tasks are complementary and can help each other in learning a better representation.
\modelname{} can seamlessly switch between FITB (id prediction) and full caption generation.

\begin{table}[t]
\centering
\small
\tabcolsep=0.16cm
\begin{tabular}{l l l cccc}
\toprule
& Captions                               & Method       & C         & M         & S          & iS         \\
\midrule
1 & \multirow{3}{*}{Fill-in \cite{fillin}}                 & Same id       & 7.03          & 9.41           & 9.01           & 0.591          \\
2 & & All diff ids & 7             & 9.11           & 12.98          & 0.202          \\ 
3 &     & FillIn & 7.77          & 10.68          & -              & -              \\ 
\midrule
4 & \multirow{3}{*}{\modelname} & Same id       & 8.44          & 10.9           & 9.26           & 0.687          \\ 
5 &    & All diff ids & 8.74          & 11.01          & 13.09          & 0.264          \\ 
6 &    & \modelname{} (Joint)   & \textbf{9.09} & \textbf{12.47} & \textbf{13.30} & \textbf{0.788} \\ 
\bottomrule
\end{tabular}
\vspace{-3mm}
\caption{We evaluate performance of id-aware captioning against~\cite{fillin}, showing improvements across all metrics.
Captioning metrics are C=CIDEr, M=METEOR, iS=\ispice, S=SPICE.}
\label{tab:sota_cap}
\vspace{-2mm}
\end{table}

\begin{table}[t]
\small
\tabcolsep=0.14cm
\centering
\begin{tabular}{l cccc c}
\toprule
\multirow{2}{*}{Method} & \multicolumn{4}{c}{Captioning metrics} & FITB \\
& C & M & S & iS & Class Acc.\\
\midrule
MICap & \textbf{9.09} & \textbf{12.47} & \textbf{13.30} & \textbf{0.788} & \textbf{70.01} \\
\midrule
T5 only CLIP & 4.9 & 8.5 & 7.1 & 0.755 & - \\
T5 all features & 4.5 & 7.9 & 6.8 & 0.723 & - \\
GPT2 only CLIP & 3.6 & 8.7 & 10.7 & 0.640 & - \\
GPT2 all features & 4.4 & 8.9 & 9.2 & 0.595 & - \\
\bottomrule
\end{tabular}
\vspace{-2mm}
\caption{Experiments showing MICap outperforms foundational models T5-Base~\cite{t5} and GPT2~\cite{gpt2} adapted/fine-tuned for id-aware captioning on the same LSMDC dataset.}
\label{tab:vlm_cap}
\vspace{-5mm}
\end{table}

\paragraph{SotA comparison for captioning.}
We compare against the two-stage baseline~\cite{fillin}, while \modelname{} predicts the captions and identities in a single stage.
\cref{tab:sota_cap} shows that we improve over~\cite{fillin} across all metrics.

\paragraph{\modelname's captions are better.}
We disentangle identity prediction from caption generation by replacing all person id labels by the same id or all different ids.
This allows us to evaluate captioning performance, independent of identity prediction.
We are pleased that our simple encoder-decoder approach outperforms a complex adversarial multi-sentence captioning approach~\cite{adversarial} used in stage 1 of~\cite{fillin}.
\cref{tab:sota_cap} R1 \vs~R4, CIDEr goes up from 7.03 to 8.44, and METEOR 9.41 to 10.9.
Similar improvements hold for R2 \vs~R5.

\paragraph{Comparison to VLMs.}
\cref{tab:vlm_cap} shows that MICap outperforms adaptations of  T5 (an encoder-decoder framework) and GPT-2 (QFormer prefix tokens like CLIPCap~\cite{clipcap} or BLIP2~\cite{blip2}), fine-tuned for the id-aware captioning task.
We suspect that integrating many diverse visual tokens is not trivial for VLMs, resulting in comparable performance when using ``only CLIP'' or ``all features''.

\paragraph{Id-aware metric.}
\ispice{} is a challenging metric as it multiplies two F1 scores that penalize when the number of identities are mismatched or tuples incorrect.
\cref{tab:sota_cap} shows that \ispice{} changes dramatically when using the same id or all different ids.
We hope that this metric will inspire future works in this direction of identity-aware captioning.

\paragraph{Attention patterns}
of \modelname's decoder reveal interesting insights.
For the task of full captioning, we see that tokens that produce id labels cross-attend more to the face tokens (from memory) while normal word tokens cross-attend to CLIP features.
We also analyze the attention patterns in FITB and observe that the model attends to the same clusters when predicting the  same labels and also attends to face detections across the videoset (not restricted to faces in a single video).
Please refer to the supplement for details.

\paragraph{A qualitative example}
is shown in \cref{fig:qualitative_eval}.
We observe that \modelname{} does a decent job at generating captions (although it is unable to use a rich vocabulary - \textit{smiles} instead of \textit{beams cheerily}).
The challenges of caption evaluation are also clear in the last clip.
Several more examples for both tasks are shown in the supplement.

\section{Conclusion}
\label{sec:conclusion}

We proposed a new paradigm for identity-aware movie caption generation.
As opposed to the two-stage approach of first captioning with anonymized names and then filling in the identities, we proposed a single-stage method that combines the two tasks via an encoder-decoder sequence-to-sequence generation framework, that can seamlessly switch between
(i)~full caption generation with identities, or (ii)~predict the identities given a caption with anonymized names.
We showed that a single auto-regressive model benefits both tasks and shows positive transfer, leading to state-of-the-art performance on the LSMDC challenge.
We also proposed an identity-aware captioning metric, \ispice, that is sensitive to subtle perturbations in identity and robustly evaluates captions.

{\small
\paragraph{Acknowledgments.}
The project was supported by funding from SERB SRG/2023/002544.
We thank the Bank of Baroda for partial travel support.
We thank Amit Pandey for assisting in early discussions.
Makarand Tapaswi thanks support from
Google India Faculty Award and
Naveen Reddy Desanur from Sensara.
}

{
\small
\bibliographystyle{ieeenat_fullname}
\bibliography{bib/longstrings,main}

\begin{thebibliography}{62}
\providecommand{\natexlab}[1]{#1}
\providecommand{\url}[1]{\texttt{#1}}
\expandafter\ifx\csname urlstyle\endcsname\relax
  \providecommand{\doi}[1]{doi: #1}\else
  \providecommand{\doi}{doi: \begingroup \urlstyle{rm}\Url}\fi

\bibitem[Anderson et~al.(2016)Anderson, Fernando, Johnson, and Gould]{spice}
Peter Anderson, Basura Fernando, Mark Johnson, and Stephen Gould.
\newblock {SPICE: Semantic Propositional Image Caption Evaluation}.
\newblock In \emph{European Conference on Computer Vision (ECCV)}, 2016.

\bibitem[Brown et~al.(2019)Brown, Albanie, Liu, Nagrani, and Zisserman]{brown19lsmdc}
Andrew Brown, Samuel Albanie, Yang Liu, Arsha Nagrani, and Andrew Zisserman.
\newblock {LSMDC v2 Challenge presentation}.
\newblock In \emph{3rd Workshop on Closing the Loop Between Vision and Language}, 2019.

\bibitem[Brown et~al.(2021)Brown, Kalogeiton, and Zisserman]{brown2021cluster}
Andrew Brown, Vicky Kalogeiton, and Andrew Zisserman.
\newblock {Face, Body, Voice: Video Person-Clustering with Multiple Modalities}.
\newblock In \emph{International Conference on Computer Vision Workshops (ICCVW)}, 2021.

\bibitem[Carreira and Zisserman(2017)]{i3d}
Joao Carreira and Andrew Zisserman.
\newblock {Quo Vadis, Action Recognition? A New Model and the Kinetics Dataset}.
\newblock In \emph{Conference on Computer Vision and Pattern Recognition (CVPR)}, 2017.

\bibitem[Chadha et~al.(2021)Chadha, Arora, and Kaloty]{Chadha2020iPerceive}
Aman Chadha, Gurneet Arora, and Navpreet Kaloty.
\newblock {{i}{P}erceive: Applying Common-Sense Reasoning to Multi-Modal Dense Video Captioning and Video Question Answering}.
\newblock In \emph{Winter Conference on Applications of Computer Vision (WACV)}, 2021.

\bibitem[Chan et~al.(2023)Chan, Petryk, Gonzalez, and Darrell]{clair}
David~M Chan, Suzanne Petryk, Joseph~E Gonzalez, and Trevor Darrell.
\newblock {CLAIR: Evaluating Image Captions with Large Language Models}.
\newblock In \emph{Empirical Methods in Natural Language Processing (EMNLP)}, 2023.

\bibitem[Chen and Jiang(2021)]{chen2021towards}
Shaoxiang Chen and Yu-Gang Jiang.
\newblock {Towards bridging event captioner and sentence localizer for weakly supervised dense event captioning}.
\newblock In \emph{Conference on Computer Vision and Pattern Recognition (CVPR)}, 2021.

\bibitem[Deng et~al.(2021)Deng, Chen, Chen, He, and Wu]{deng2021sketch}
Chaorui Deng, Shizhe Chen, Da Chen, Yuan He, and Qi Wu.
\newblock {Sketch, ground, and refine: Top-down dense video captioning}.
\newblock In \emph{Conference on Computer Vision and Pattern Recognition (CVPR)}, 2021.

\bibitem[Deng et~al.(2019)Deng, Guo, Xue, and Zafeiriou]{arcface}
Jiankang Deng, Jia Guo, Niannan Xue, and Stefanos Zafeiriou.
\newblock {ArcFace: Additive Angular Margin Loss for Deep Face Recognition}.
\newblock In \emph{Conference on Computer Vision and Pattern Recognition (CVPR)}, 2019.

\bibitem[{Deng, Jiankang and Guo, Jia and Ververas, Evangelos and Kotsia, Irene and Zafeiriou, Stefanos}(2020)]{retina}
{Deng, Jiankang and Guo, Jia and Ververas, Evangelos and Kotsia, Irene and Zafeiriou, Stefanos}.
\newblock Retinaface: Single-shot multi-level face localisation in the wild.
\newblock In \emph{Conference on Computer Vision and Pattern Recognition (CVPR)}, 2020.

\bibitem[Denkowski and Lavie(2014)]{meteor}
Michael Denkowski and Alon Lavie.
\newblock {Meteor Universal: Language Specific Translation Evaluation for Any Target Language}.
\newblock In \emph{European Chapter of the Association for Computational Linguistics (EACL)}, 2014.

\bibitem[Donahue et~al.(2015)Donahue, Anne~Hendricks, Guadarrama, Rohrbach, Venugopalan, Saenko, and Darrell]{lrcn}
Jeffrey Donahue, Lisa Anne~Hendricks, Sergio Guadarrama, Marcus Rohrbach, Subhashini Venugopalan, Kate Saenko, and Trevor Darrell.
\newblock {Long-term recurrent convolutional networks for visual recognition and description}.
\newblock In \emph{Conference on Computer Vision and Pattern Recognition (CVPR)}, 2015.

\bibitem[Han et~al.(2023{\natexlab{a}})Han, Bain, Nagrani, Varol, Xie, and Zisserman]{autoad}
Tengda Han, Max Bain, Arsha Nagrani, G{\"u}l Varol, Weidi Xie, and Andrew Zisserman.
\newblock {AutoAD: Movie Description in Context}.
\newblock In \emph{Conference on Computer Vision and Pattern Recognition (CVPR)}, 2023{\natexlab{a}}.

\bibitem[Han et~al.(2023{\natexlab{b}})Han, Bain, Nagrani, Varol, Xie, and Zisserman]{autoad2}
Tengda Han, Max Bain, Arsha Nagrani, Gul Varol, Weidi Xie, and Andrew Zisserman.
\newblock {AutoAD II: The Sequel-Who, When, and What in Movie Audio Description}.
\newblock In \emph{International Conference on Computer Vision (ICCV)}, 2023{\natexlab{b}}.

\bibitem[Hessel et~al.(2021)Hessel, Holtzman, Forbes, Bras, and Choi]{clipscore}
Jack Hessel, Ari Holtzman, Maxwell Forbes, Ronan~Le Bras, and Yejin Choi.
\newblock {CLIPScore: A Reference-free Evaluation Metric for Image Captioning}.
\newblock In \emph{Empirical Methods in Natural Language Processing (EMNLP)}, 2021.

\bibitem[Johnson et~al.(2015)Johnson, Krishna, Stark, Li, Shamma, Bernstein, and Fei-Fei]{imagescene}
Justin Johnson, Ranjay Krishna, Michael Stark, Li-Jia Li, David Shamma, Michael Bernstein, and Li Fei-Fei.
\newblock {Image Retrieval using Scene Graphs}.
\newblock In \emph{Conference on Computer Vision and Pattern Recognition (CVPR)}, 2015.

\bibitem[Khan et~al.(2022)Khan, Jawahar, and Tapaswi]{GVSR}
Zeeshan Khan, CV Jawahar, and Makarand Tapaswi.
\newblock {Grounded Video Situation Recognition}.
\newblock In \emph{Advances in Neural Information Processing Systems (NeurIPS)}, 2022.

\bibitem[Krishna et~al.(2017)Krishna, Hata, Ren, Fei-Fei, and Carlos~Niebles]{krishna2017dense}
Ranjay Krishna, Kenji Hata, Frederic Ren, Li Fei-Fei, and Juan Carlos~Niebles.
\newblock {Dense-captioning events in videos}.
\newblock In \emph{International Conference on Computer Vision (ICCV)}, 2017.

\bibitem[Li et~al.(2023)Li, Li, Savarese, and Hoi]{blip2}
Junnan Li, Dongxu Li, Silvio Savarese, and Steven Hoi.
\newblock {BLIP-2: Bootstrapping Language-Image Pre-training with Frozen Image Encoders and Large Language Models}.
\newblock In \emph{International Conference on Machine Learning (ICML)}, 2023.

\bibitem[Li et~al.(2018)Li, Yao, Pan, Chao, and Mei]{li2018jointly}
Yehao Li, Ting Yao, Yingwei Pan, Hongyang Chao, and Tao Mei.
\newblock {Jointly localizing and describing events for dense video captioning}.
\newblock In \emph{Conference on Computer Vision and Pattern Recognition (CVPR)}, 2018.

\bibitem[Lin(2004)]{rouge}
Chin-Yew Lin.
\newblock {ROUGE: A Package for Automatic Evaluation of Summaries}.
\newblock In \emph{Workshop on Text Summarization Branches Out (WAS)}, 2004.

\bibitem[Lin et~al.(2022)Lin, Li, Lin, Ahmed, Gan, Liu, Lu, and Wang]{swinbert}
Kevin Lin, Linjie Li, Chung-Ching Lin, Faisal Ahmed, Zhe Gan, Zicheng Liu, Yumao Lu, and Lijuan Wang.
\newblock {Swinbert: End-to-end transformers with sparse attention for video captioning}.
\newblock In \emph{Conference on Computer Vision and Pattern Recognition (CVPR)}, 2022.

\bibitem[Luo et~al.(2020)Luo, Ji, Shi, Huang, Duan, Li, Li, Bharti, and Zhou]{univl}
Huaishao Luo, Lei Ji, Botian Shi, Haoyang Huang, Nan Duan, Tianrui Li, Jason Li, Taroon Bharti, and Ming Zhou.
\newblock {UniVL: A Unified Video and Language Pre-training Model for Multimodal Understanding and Generation}.
\newblock \emph{arXiv preprint arXiv:2002.06353}, 2020.

\bibitem[Mokady et~al.(2021)Mokady, Hertz, and Bermano]{clipcap}
Ron Mokady, Amir Hertz, and Amit~H. Bermano.
\newblock {ClipCap: CLIP Prefix for Image Captioning}.
\newblock \emph{arXiv preprint 2111.09734}, 2021.

\bibitem[Mun et~al.(2019)Mun, Yang, Ren, Xu, and Han]{mun2019streamlined}
Jonghwan Mun, Linjie Yang, Zhou Ren, Ning Xu, and Bohyung Han.
\newblock {Streamlined dense video captioning}.
\newblock In \emph{Conference on Computer Vision and Pattern Recognition (CVPR)}, 2019.

\bibitem[Nagrani and Zisserman(2017)]{sherlock_personid}
Arsha Nagrani and Andrew Zisserman.
\newblock {From Benedict Cumberbatch to Sherlock Holmes: Character Identification in TV series without a Script}.
\newblock In \emph{British Machine Vision Conference (BMVC)}, 2017.

\bibitem[Papineni et~al.(2002)Papineni, Roukos, Ward, and Zhu]{bleu}
Kishore Papineni, Salim Roukos, Todd Ward, and Wei-Jing Zhu.
\newblock {BLEU: a method for automatic evaluation of machine translation}.
\newblock In \emph{Association of Computational Linguistics (ACL)}, 2002.

\bibitem[Park et~al.(2019)Park, Rohrbach, Darrell, and Rohrbach]{adversarial}
Jae~Sung Park, Marcus Rohrbach, Trevor Darrell, and Anna Rohrbach.
\newblock {Adversarial inference for multi-sentence video description}.
\newblock In \emph{Conference on Computer Vision and Pattern Recognition (CVPR)}, 2019.

\bibitem[Park et~al.(2020)Park, Darrell, and Rohrbach]{fillin}
Jae~Sung Park, Trevor Darrell, and Anna Rohrbach.
\newblock {Identity-aware multi-sentence video description}.
\newblock In \emph{European Conference on Computer Vision (ECCV)}, 2020.

\bibitem[Pini et~al.(2017)Pini, Cornia, Baraldi, and Cucchiara]{mvad}
Stefano Pini, Marcella Cornia, Lorenzo Baraldi, and Rita Cucchiara.
\newblock {Towards video captioning with naming: a novel dataset and a multi-modal approach}.
\newblock In \emph{International Conference on Image Analysis and Processing (ICIAP)}, 2017.

\bibitem[Pini et~al.(2019)Pini, Cornia, Bolelli, Baraldi, and Cucchiara]{mvadnames}
Stefano Pini, Marcella Cornia, Federico Bolelli, Lorenzo Baraldi, and Rita Cucchiara.
\newblock {M-VAD names: a dataset for video captioning with naming}.
\newblock \emph{Multimedia Tools and Applications (MTAP)}, 78:\penalty0 14007--14027, 2019.

\bibitem[Radford et~al.(2019)Radford, Wu, Child, Luan, Amodei, and Sutskever]{gpt2}
Alec Radford, Jeff Wu, Rewon Child, David Luan, Dario Amodei, and Ilya Sutskever.
\newblock {Language Models are Unsupervised Multitask Learners}.
\newblock 2019.

\bibitem[Radford et~al.(2021)Radford, Kim, Hallacy, Ramesh, Goh, Agarwal, Sastry, Askell, Mishkin, Clark, et~al.]{clip}
Alec Radford, Jong~Wook Kim, Chris Hallacy, Aditya Ramesh, Gabriel Goh, Sandhini Agarwal, Girish Sastry, Amanda Askell, Pamela Mishkin, Jack Clark, et~al.
\newblock {Learning transferable visual models from natural language supervision}.
\newblock In \emph{International Conference on Machine Learning (ICML)}. PMLR, 2021.

\bibitem[Raffel et~al.(2020)Raffel, Shazeer, Roberts, Lee, Narang, Matena, Zhou, Li, and Liu]{t5}
Colin Raffel, Noam Shazeer, Adam Roberts, Katherine Lee, Sharan Narang, Michael Matena, Yanqi Zhou, Wei Li, and Peter~J. Liu.
\newblock {Exploring the Limits of Transfer Learning with a Unified Text-to-Text Transformer}.
\newblock \emph{Journal of Machine Learning Research (JMLR)}, 21:\penalty0 1--67, 2020.

\bibitem[Rahman et~al.(2019)Rahman, Xu, and Sigal]{rahman2019watch}
Tanzila Rahman, Bicheng Xu, and Leonid Sigal.
\newblock {Watch, listen and tell: Multi-modal weakly supervised dense event captioning}.
\newblock In \emph{International Conference on Computer Vision (ICCV)}, 2019.

\bibitem[Rohrbach et~al.(2014)Rohrbach, Rohrbach, Qiu, Friedrich, Pinkal, and Schiele]{rohrbachcoherent}
Anna Rohrbach, Marcus Rohrbach, Wei Qiu, Annemarie Friedrich, Manfred Pinkal, and Bernt Schiele.
\newblock {Coherent multi-sentence video description with variable level of detail}.
\newblock In \emph{German Conference on Pattern Recognition (GCPR)}, 2014.

\bibitem[Rohrbach et~al.(2015)Rohrbach, Rohrbach, Tandon, and Schiele]{mpiimd}
Anna Rohrbach, Marcus Rohrbach, Niket Tandon, and Bernt Schiele.
\newblock {A Dataset for Movie Description}.
\newblock In \emph{Conference on Computer Vision and Pattern Recognition (CVPR)}, 2015.

\bibitem[Rohrbach et~al.(2017)Rohrbach, Torabi, Rohrbach, Tandon, Pal, Larochelle, Courville, and Schiele]{lsmdc}
Anna Rohrbach, Atousa Torabi, Marcus Rohrbach, Niket Tandon, Christopher Pal, Hugo Larochelle, Aaron Courville, and Bernt Schiele.
\newblock {Movie description}.
\newblock \emph{International Journal of Computer Vision (IJCV)}, 123:\penalty0 94--120, 2017.

\bibitem[Sadhu et~al.(2021)Sadhu, Gupta, Yatskar, Nevatia, and Kembhavi]{vidsitu}
Arka Sadhu, Tanmay Gupta, Mark Yatskar, Ram Nevatia, and Aniruddha Kembhavi.
\newblock {Visual Semantic Role Labeling for Video Understanding}.
\newblock In \emph{Conference on Computer Vision and Pattern Recognition (CVPR)}, 2021.

\bibitem[Schroff et~al.(2015)Schroff, Kalenichenko, and Philbin]{facenet}
Florian Schroff, Dmitry Kalenichenko, and James Philbin.
\newblock {FaceNet: A Unified Embedding for Face Recognition and Clustering}.
\newblock In \emph{Conference on Computer Vision and Pattern Recognition (CVPR)}, 2015.

\bibitem[Schuster et~al.(2015)Schuster, Krishna, Chang, Fei-Fei, and Manning]{scenegraph}
Sebastian Schuster, Ranjay Krishna, Angel Chang, Li Fei-Fei, and Christopher~D Manning.
\newblock {Generating semantically precise scene graphs from textual descriptions for improved image retrieval}.
\newblock In \emph{Fourth Workshop on Vision and Language}, 2015.

\bibitem[Seo et~al.(2022)Seo, Nagrani, Arnab, and Schmid]{seoend}
Paul~Hongsuck Seo, Arsha Nagrani, Anurag Arnab, and Cordelia Schmid.
\newblock {End-to-end generative pretraining for multimodal video captioning}.
\newblock In \emph{Conference on Computer Vision and Pattern Recognition (CVPR)}, 2022.

\bibitem[Shen et~al.(2017)Shen, Li, Su, Li, Chen, Jiang, and Xue]{shen2017weakly}
Zhiqiang Shen, Jianguo Li, Zhou Su, Minjun Li, Yurong Chen, Yu-Gang Jiang, and Xiangyang Xue.
\newblock {Weakly supervised dense video captioning}.
\newblock In \emph{Conference on Computer Vision and Pattern Recognition (CVPR)}, 2017.

\bibitem[Shi et~al.(2019)Shi, Ji, Liang, Duan, Chen, Niu, and Zhou]{shi2019dense}
Botian Shi, Lei Ji, Yaobo Liang, Nan Duan, Peng Chen, Zhendong Niu, and Ming Zhou.
\newblock {Dense procedure captioning in narrated instructional videos}.
\newblock In \emph{Association of Computational Linguistics (ACL)}, 2019.

\bibitem[Shin et~al.(2016)Shin, Ohnishi, and Harada]{shin2016beyond}
Andrew Shin, Katsunori Ohnishi, and Tatsuya Harada.
\newblock {Beyond caption to narrative: Video captioning with multiple sentences}.
\newblock In \emph{International Conference on Image Processing (ICIP)}, 2016.

\bibitem[{Soldan, Mattia and Pardo, Alejandro and Alc\'azar, Juan Le\'on and Caba, Fabian and Zhao, Chen and Giancola, Silvio and Ghanem, Bernard}(2022)]{MAD}
{Soldan, Mattia and Pardo, Alejandro and Alc\'azar, Juan Le\'on and Caba, Fabian and Zhao, Chen and Giancola, Silvio and Ghanem, Bernard}.
\newblock {MAD: A Scalable Dataset for Language Grounding in Videos From Movie Audio Descriptions}.
\newblock In \emph{Conference on Computer Vision and Pattern Recognition (CVPR)}, 2022.

\bibitem[Tapaswi et~al.(2012)Tapaswi, Bäuml, and Stiefelhagen]{knockknock}
Makarand Tapaswi, Martin Bäuml, and Rainer Stiefelhagen.
\newblock {``Knock! Knock! Who is it?" Probabilistic Person Identification in TV series}.
\newblock In \emph{Conference on Computer Vision and Pattern Recognition (CVPR)}, 2012.

\bibitem[Tapaswi et~al.(2019)Tapaswi, Law, and Fidler]{ballclustering}
Makarand Tapaswi, Marc~T. Law, and Sanja Fidler.
\newblock {Video Face Clustering with Unknown Number of Clusters}.
\newblock In \emph{International Conference on Computer Vision (ICCV)}, 2019.

\bibitem[Vaswani et~al.(2017)Vaswani, Shazeer, Parmar, Uszkoreit, Jones, Gomez, Kaiser, and Polosukhin]{transformer}
Ashish Vaswani, Noam Shazeer, Niki Parmar, Jakob Uszkoreit, Llion Jones, Aidan~N Gomez, {\L}ukasz Kaiser, and Illia Polosukhin.
\newblock {Attention is All You Need}.
\newblock In \emph{Advances in Neural Information Processing Systems (NeurIPS)}, 2017.

\bibitem[Vedantam et~al.(2015)Vedantam, Zitnick, and Parikh]{cider}
Ramakrishna Vedantam, C.~Lawrence Zitnick, and Devi Parikh.
\newblock {CIDEr: Consensus-based image description evaluation}.
\newblock In \emph{Conference on Computer Vision and Pattern Recognition (CVPR)}, 2015.

\bibitem[Venugopalan et~al.(2015{\natexlab{a}})Venugopalan, Rohrbach, Donahue, Mooney, Darrell, and Saenko]{s2svt}
Subhashini Venugopalan, Marcus Rohrbach, Jeffrey Donahue, Raymond Mooney, Trevor Darrell, and Kate Saenko.
\newblock {Sequence to sequence-video to text}.
\newblock In \emph{International Conference on Computer Vision (ICCV)}, 2015{\natexlab{a}}.

\bibitem[Venugopalan et~al.(2015{\natexlab{b}})Venugopalan, Xu, Donahue, Rohrbach, Mooney, and Saenko]{tvnl}
Subhashini Venugopalan, Huijuan Xu, Jeff Donahue, Marcus Rohrbach, Raymond Mooney, and Kate Saenko.
\newblock {Translating Videos to Natural Language Using Deep Recurrent Neural Networks}.
\newblock In \emph{North American Chapter of the Association for Computational Linguistics: Human Language Technologies}, 2015{\natexlab{b}}.

\bibitem[Wang et~al.(2018)Wang, Jiang, Ma, Liu, and Xu]{wang2018bidirectional}
Jingwen Wang, Wenhao Jiang, Lin Ma, Wei Liu, and Yong Xu.
\newblock {Bidirectional Attentive Fusion with Context Gating for Dense Video Captioning}.
\newblock In \emph{Conference on Computer Vision and Pattern Recognition (CVPR)}, 2018.

\bibitem[Wang et~al.(2020)Wang, Zheng, Yu, Tian, and Hu]{wang2020event}
Teng Wang, Huicheng Zheng, Mingjing Yu, Qian Tian, and Haifeng Hu.
\newblock {Event-centric hierarchical representation for dense video captioning}.
\newblock \emph{IEEE Transactions on Circuits and Systems for Video Technology}, 31\penalty0 (5):\penalty0 1890--1900, 2020.

\bibitem[Wang et~al.(2021)Wang, Zhang, Lu, Zheng, Cheng, and Luo]{wang2021end}
Teng Wang, Ruimao Zhang, Zhichao Lu, Feng Zheng, Ran Cheng, and Ping Luo.
\newblock {End-to-end Dense Video Captioning with Parallel Decoding}.
\newblock In \emph{International Conference on Computer Vision (ICCV)}, 2021.

\bibitem[Yang et~al.(2023)Yang, Nagrani, Seo, Miech, Pont-Tuset, Laptev, Sivic, and Schmid]{vid2seq}
Antoine Yang, Arsha Nagrani, Paul~Hongsuck Seo, Antoine Miech, Jordi Pont-Tuset, Ivan Laptev, Josef Sivic, and Cordelia Schmid.
\newblock {Vid2seq: Large-scale pretraining of a visual language model for dense video captioning}.
\newblock In \emph{Conference on Computer Vision and Pattern Recognition (CVPR)}, 2023.

\bibitem[Yu et~al.(2016)Yu, Wang, Huang, Yang, and Xu]{yu2016video}
Haonan Yu, Jiang Wang, Zhiheng Huang, Yi Yang, and Wei Xu.
\newblock {Video paragraph captioning using hierarchical recurrent neural networks}.
\newblock In \emph{Conference on Computer Vision and Pattern Recognition (CVPR)}, 2016.

\bibitem[Yu et~al.(2017)Yu, Ko, Choi, and Kim]{eterqa}
Youngjae Yu, Hyungjin Ko, Jongwook Choi, and Gunhee Kim.
\newblock {End-to-end concept word detection for video captioning, retrieval, and question answering}.
\newblock In \emph{Conference on Computer Vision and Pattern Recognition (CVPR)}, 2017.

\bibitem[Yu et~al.(2019)Yu, Chung, Kim, Yun, and Kim]{yu19lsmdc}
Youngjae Yu, Jiwan Chung, Jongseok Kim, Heeseung Yun, and Gunhee Kim.
\newblock {LSMDC v2 Challenge presentation}.
\newblock 2019.

\bibitem[Yu et~al.(2020)Yu, Kim, Yun, Jiwan, and Kim]{cisin}
Youngjae Yu, Jongseok Kim, Heeseung Yun, Chung Jiwan, and Gunhee Kim.
\newblock {{Character Grounding and Re-Identification inStory of Videos and Text Descriptions}}.
\newblock In \emph{European Conference on Computer Vision (ECCV)}, 2020.

\bibitem[Zhang* et~al.(2020)Zhang*, Kishore*, Wu*, Weinberger, and Artzi]{bertscore}
Tianyi Zhang*, Varsha Kishore*, Felix Wu*, Kilian~Q. Weinberger, and Yoav Artzi.
\newblock {BERTScore: Evaluating Text Generation with BERT}.
\newblock In \emph{International Conference on Learning Representations (ICLR)}, 2020.

\bibitem[Zhou et~al.(2018)Zhou, Zhou, Corso, Socher, and Xiong]{zhou2018end}
Luowei Zhou, Yingbo Zhou, Jason~J Corso, Richard Socher, and Caiming Xiong.
\newblock {End-to-end dense video captioning with masked transformer}.
\newblock In \emph{Conference on Computer Vision and Pattern Recognition (CVPR)}, 2018.

\end{thebibliography}
}

\appendix
\clearpage

\begin{figure*}
\centering
\includegraphics[width=0.9\linewidth]{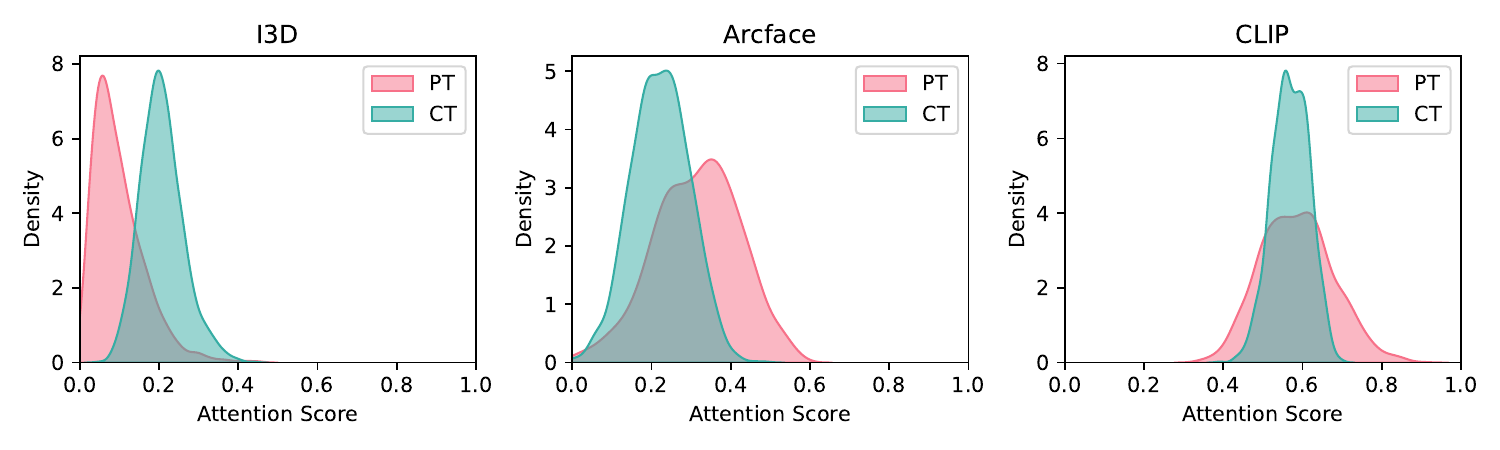}
\vspace{-3mm}
\captionof{figure}{Cross-attention scores density plots for the id-aware captioning task.
We group decoder output tokens into two types: person id label tokens (PT), and caption tokens that represent other words (CT).
Attention scores are grouped across the three input visual features capturing actions (I3D, left), faces (Arcface, middle), and semantic content (CLIP, right).
Please refer to \cref{subsec:supp_attn_idwarecap} for a discussion.}
\label{fig:supp_idcap_visualmod}
\vspace{6mm}
\end{figure*}

\section*{Appendix}

\noindent
We present additional insights and results in the supplementary material.
In \cref{sec:supp_attention}, we highlight how our auto-regressive Transformer decoder attends to various memory features.
For the id-aware captioning task, we show the relative importance of the 3 visual features, while
for the Fill-in-the-blanks (FITB) task, we highlight how our model attends to correct face clusters.
Next, in \cref{sec:supp_qualitative}, we show qualitative results for both tasks, FITB and id-aware captioning.
We also illustrate how our new identity-aware metric, \ispice{}, is calculated on some examples.
Finally, we end with discussion of some limitations in \cref{sec:supp_limitations}.

\section{Analyzing Model Attention}
\label{sec:supp_attention}

In this section, we visualize and discuss the attention scores from \modelname's auto-regressive Transformer decoder.
In particular, we focus on the cross-attention scores of the last layer as they reveal interesting insights about the features that the captioning model uses.
Throughout this section, we analyze \modelname{} trained jointly on id-aware captioning and FITB.
All attention scores are obtained in inference mode.

\subsection{Attention Patterns in Id-aware Captioning}
\label{subsec:supp_attn_idwarecap}

In id-aware full captioning, for a particular videoset $\MN = \{V_i\}_{i=1}^N$, we first encode the videos to obtain memory tokens $M$ and pass them through a Transformer decoder auto-regressively to generate one token (word) at a time.
If we consider that the number of tokens in the predicted captionset is $L$, we can compute a matrix of cross-attention scores $\alpha = L \times |M|$, where $|M|$ is the number of tokens in the decoder memory.
Note, while we use multi-head attention, scores over the heads are averaged obtain $\alpha$.

We split the $L$ tokens into 2 groups:
(i)~one group consists of person id label predictions or \emph{person tokens} (PT); and
(ii)~the other group consists of all other tokens referred to as \emph{caption tokens} (CT).
For visualization, we sum over the attention scores for each of the token types (id labels and text) and convert our attention map to a matrix of $2 \times |M|$.

Next, we also group the memory tokens into 3 types of visual features used in our work: action (I3D), face (Arcface), and semantic features (CLIP).
Thus, we obtain a $2 \times 3$ matrix of cross-attention scores for each sample.

\paragraph{Results.}
We compute attention scores over all samples of the validation set and plot them as a probability density function in \cref{fig:supp_idcap_visualmod}.
PT (red) and CT (green) represent the person and caption tokens respectively.
We observe that:
(i)~The model relies on CLIP features to predict captions (depicted by the overall high attention scores from 0.5-0.7).
(ii)~When predicting person tokens (PT) of the identity-aware captions, the model tends to look at face features (0.1-0.6) more than when predicting caption tokens (0-0.4).
(iii)~Finally, while action features are useful for captioning, they are less useful for predicting person-id labels. This is expected as action recognition is an identity-agnostic task.

\begin{figure*}[t]
\centering
\small
\textbf{Captionset 1}: Someone watches the aliens draw closer. 
\makeblank{} sits back in the doorway clutching a radio.
\makeblank{} watches from his position several yards away.
\makeblank{} squeezes the detonator the bus blows apart. \\
\includegraphics[width=0.32\linewidth]{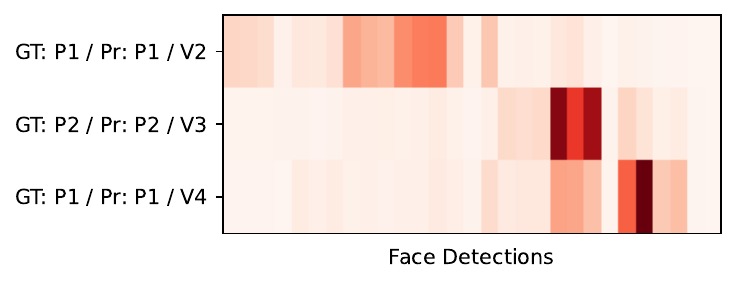}
\includegraphics[width=0.32\linewidth]{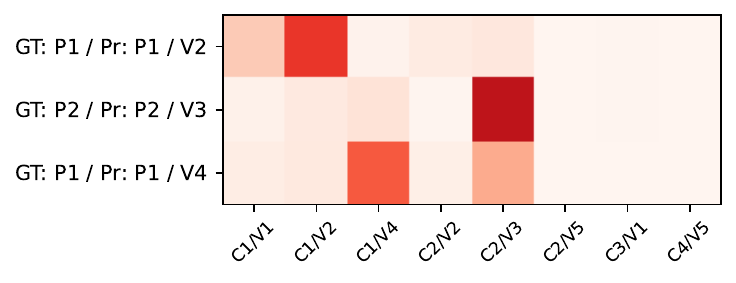}
\includegraphics[width=0.32\linewidth]{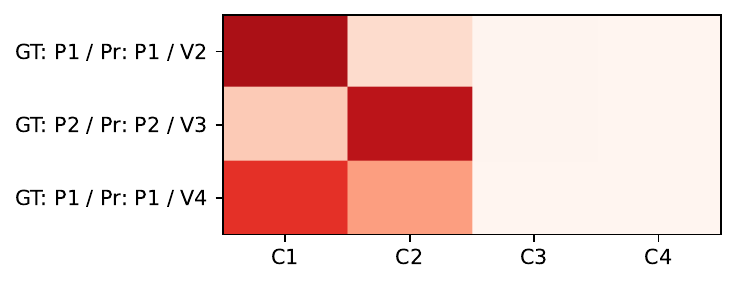} \\
\vspace{7mm}
\textbf{Captionset 2}: \makeblank{} and \makeblank{} killed their first witch.
They advance cautiously.
Suddenly \makeblank{} is thrown to the ground with a jolt.
\makeblank{} whips around a weapon poised to find \makeblank{} holding her wand to neck.
\makeblank{} begins to put the gun on the ground. \\
\includegraphics[width=0.32\linewidth]{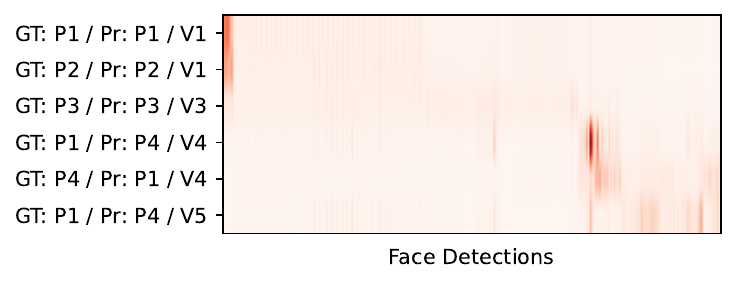}
\includegraphics[width=0.32\linewidth]{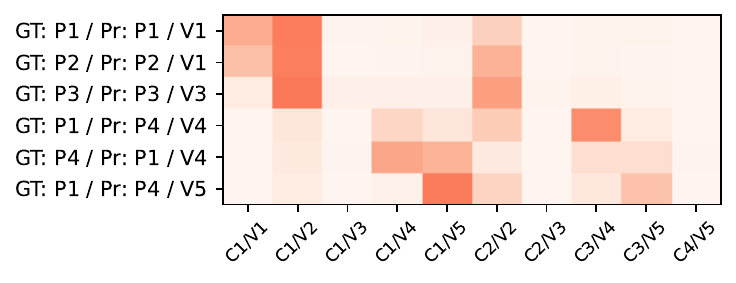}
\includegraphics[width=0.32\linewidth]{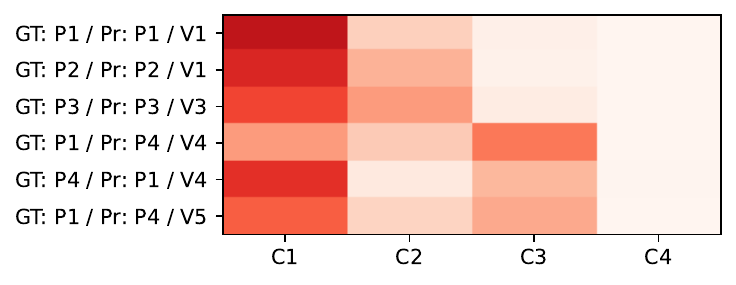} \\
\vspace{7mm}
\textbf{Captionset 3}: \makeblank{} pulls her phone from her bag and answers.
\makeblank{} frowns uncertainly.
\makeblank{} leans on a wall and slips.
\makeblank{} lowers his phone and folds it shut.
The next morning two women stroll across the street in front of apartment building. \\
\includegraphics[width=0.32\linewidth]{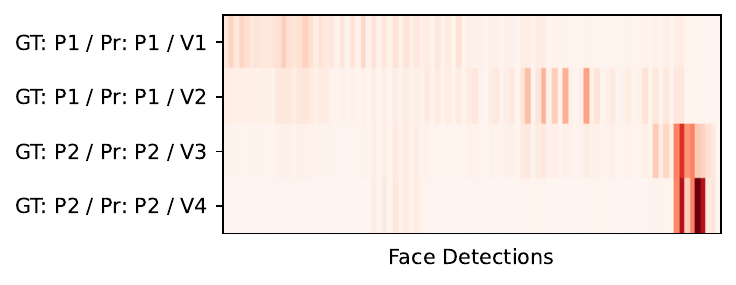}
\includegraphics[width=0.32\linewidth]{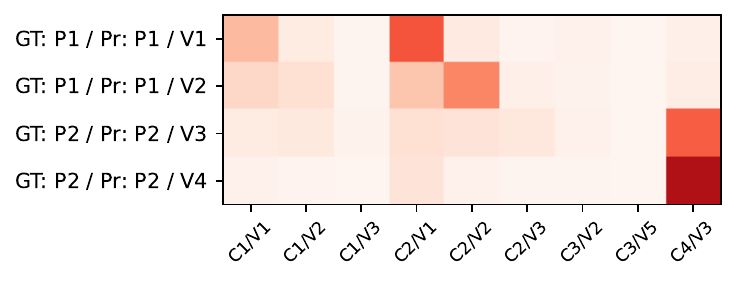}
\includegraphics[width=0.32\linewidth]{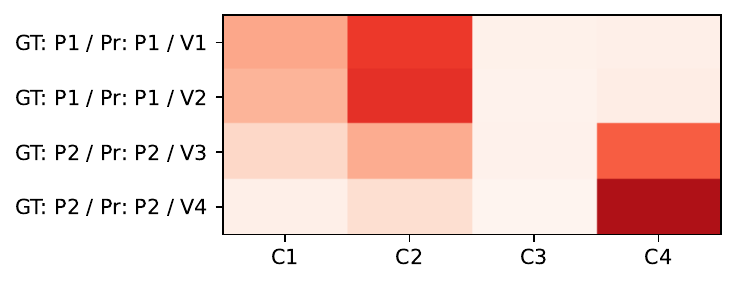} \\
\vspace{7mm}
\textbf{Captionset 4}: \makeblank{} scrutinizes his earnest face.
His eyes gleaming in the dim light.
\makeblank{} abruptly gets to his feet and heads for the door now.
\makeblank{} talks on his cell as \makeblank{} steps into the daylight silhouetted against the sunny day.
\makeblank{} faces the door frame and leans his head against it now.
In a hotel suite a woman applies makeup to \makeblank{}. \\
\includegraphics[width=0.32\linewidth]{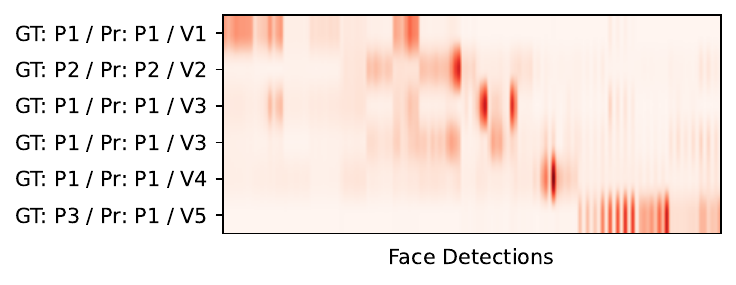}
\includegraphics[width=0.32\linewidth]{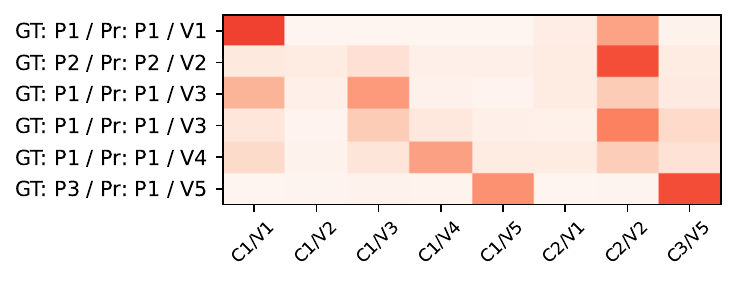}
\includegraphics[width=0.32\linewidth]{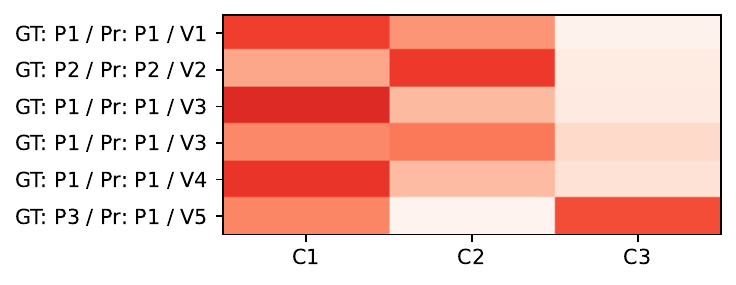} \\
\vspace{7mm}
\textbf{Captionset 5}: \makeblank{} turns and spots the brown chevy 4x4 parked on a short driveway. 
\makeblank{} approaches the vehicle cautiously across a lawn leaning over to get a view of its occupant.
The passenger side window is lowered.
\makeblank{} puts both hands on the sill and leans in with an inquisitive frown.
\makeblank{}, the asian man who in town sits with one hand clamped to the steering wheel rocking nervously and staring numbly ahead. \\
\includegraphics[width=0.32\linewidth]{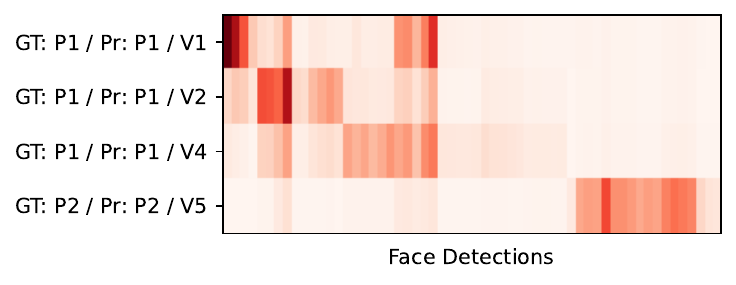}
\includegraphics[width=0.32\linewidth]{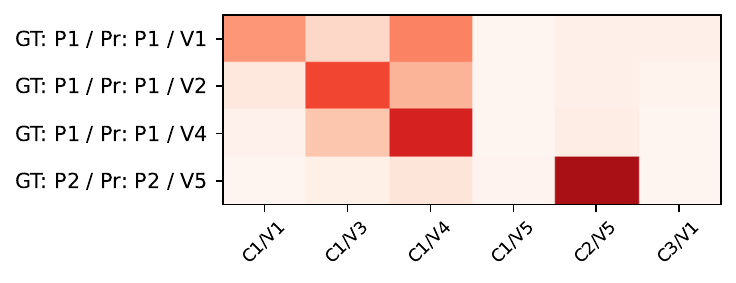}
\includegraphics[width=0.32\linewidth]{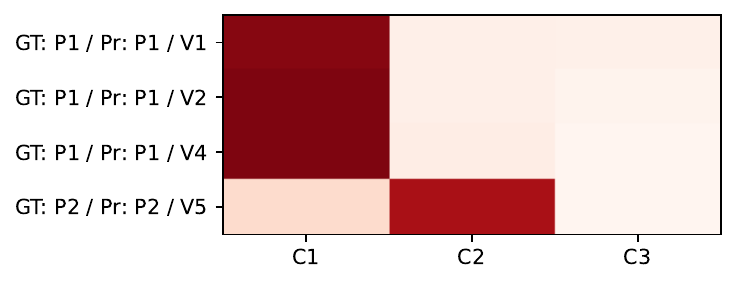} \\
\vspace{5mm}
\caption{We show 5 examples of our model's attention scores on the FITB task.
For each example (row), we show the captionset (with blanks) and the attention scores grouped in various ways.
The \textbf{left} column shows the attention score for each blank across all face detections in the video.
The \textbf{middle} column shows attention scores for face detections grouped by clusters in each video. C1/V1 indicates faces appearing in cluster 1 and video 1, while C1/V2 indicates faces of the same cluster 1 appearing in video 2.
The \textbf{right} column shows attention scores of each blank for face clusters (across videos).
For each row in the attention scores, we indicate the ground-truth (GT) and predicted (Pr) person id label and the video index (V1 .. V5) in which this blank appeared.
See \cref{subsec:supp_attnfitb} for a discussion.
}
\label{fig:supp_attn_fitb}
\end{figure*}

\subsection{Attention Patterns in FITB}
\label{subsec:supp_attnfitb}

For the FITB task, we analyze how the person id predictions attend to \textit{face features} from the decoder memory.
For a videoset $\MN = \{V_i\}_{i=1}^N$ and its corresponding captionset with blanks $\hat{\MC}$ we obtain a cross-attention map of $\alpha = |\MB| \times F$, where $|\MB|$ is the number of blanks in the captionset, and $F$ is the number of face detections across the videoset.
Each row of this matrix is normalized to sum to 1.

The attention scores and captionsets with blanks are presented in \cref{fig:supp_attn_fitb}.
In the next paragraphs, we will analyze the 3 types (columns) of the presented scores.

\paragraph{Cross-attention scores for face detections.}
In the left column of \cref{fig:supp_attn_fitb}, we visualize the attention scores directly for each face detection.
In the plot, x-axis spans time across different videos.
Our model tends to show a diagonal pattern indicating that person id label predictions tend to look at faces in the same video (facilitated through the $\bE^\text{vid}$ embeddings).
However, as seen in captionset 5, left, row 1, the model may also attend to other face detections of the same person across videos.
This highlights that being able to attend to faces across videos is useful (compared to~\cite{fillin} that only looks at faces within the same video).

\paragraph{Cross-attention scores for face clusters grouped by video index.}
Shown in the middle column of \cref{fig:supp_attn_fitb}, we group the $F$ face detections into clusters, but split them based on video index in the videoset.
For example, in captionset 1, we see that faces in cluster 1 appears across videos 1, 2, 4 (C1/V1, C1/V2, C1/V4).
This allows us to explain some of the predictions made by our model.

Please note that the face cluster index and person id labels need not match numerically.
That is, cluster 2 could be assigned the label P1 and cluster 1 the label P2.
These changes are acceptable as we only consider person id labels in a local videoset.

In cationset 3, we see that cluster 2 corresponds to the prediction P1 (first two rows) and cluster 4 (C4/V3) corresponds to person id label P2 (bottom two rows).
In the last row of captionset 3, we see that our model predicts P2 for the video id 4 correctly, while looking at cluster 4 in video 3 (C4/V3).
Previous work~\cite{fillin} is unable to use such cross-video information.

\begin{figure}[t]
\centering
\includegraphics[width=\linewidth]{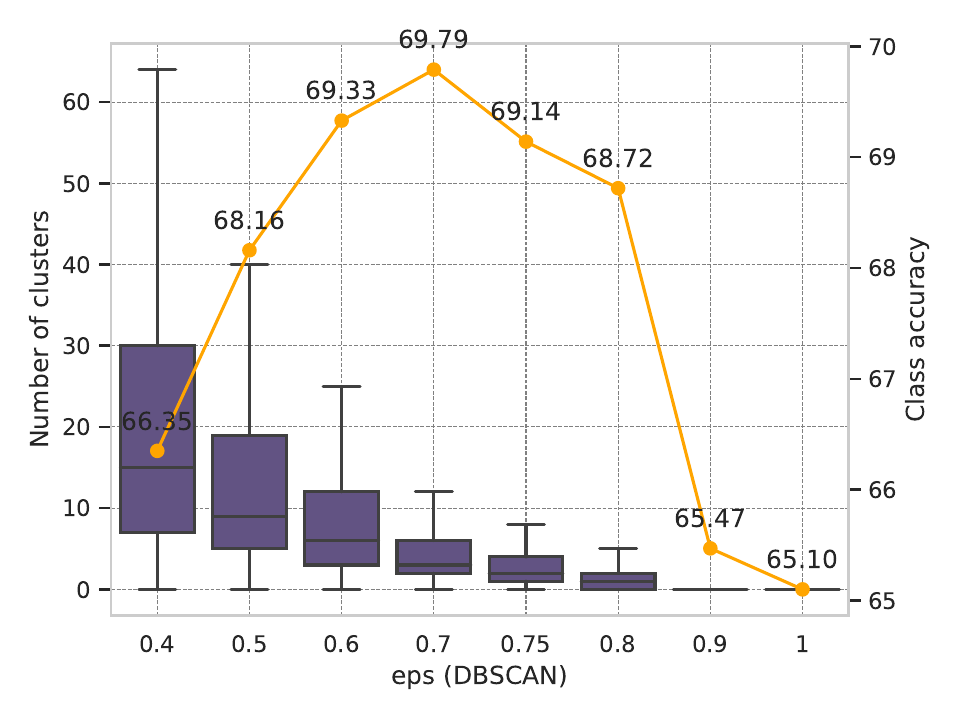}
\vspace{-5mm}
\caption{Class-accuracy for the FITB task by varying the DBSCAN eps distance threshold.
We also show a box-plot for the number of clusters created at each threshold across samples of the validation set.}
\label{fig:impact_dbscan_eps}
\vspace{-5mm}
\end{figure}

\paragraph{Cross-attention scores for clusters.}
In the right columns of \cref{fig:supp_attn_fitb}, we show attention scores directly grouped by cluster ids.
Here, the original attention map of $|\MB| \times F$ is grouped to $|\MB| \times |\MG|$, where $|\MG|$ is the number of face clusters obtained after performing DBSCAN on the $F$ face detections.

Captionset 2 is an example with multiple blanks and 4 characters. We observe that some confusion in attention scores leads to errors in the predicted person id labels.
In captionset 4, we also see 6 blanks, now with 3 characters.
In the last row, while the model wrongly predicts P1, the model does look at cluster 3 (corresponding to P3) correctly.
Captionset 1 and 2 are examples of perfect attention scores and clusters. P1 and C1, and P2 and C2 go together strongly in these examples.

\paragraph{Impact of number of clusters on FITB.}
\cref{fig:impact_dbscan_eps} shows the results on FITB class-accuracy for varying the DBSCAN epsilon parameter.
These results indicate the importance of clustering across videos and choosing an appropriate number of clusters.
Qualitatively, we adopt 0.75 as it is unlikely to merge characters incorrectly.

\begin{figure*}[p]
\centering
\raisebox{1.1cm}{\includegraphics[width=0.48\linewidth]{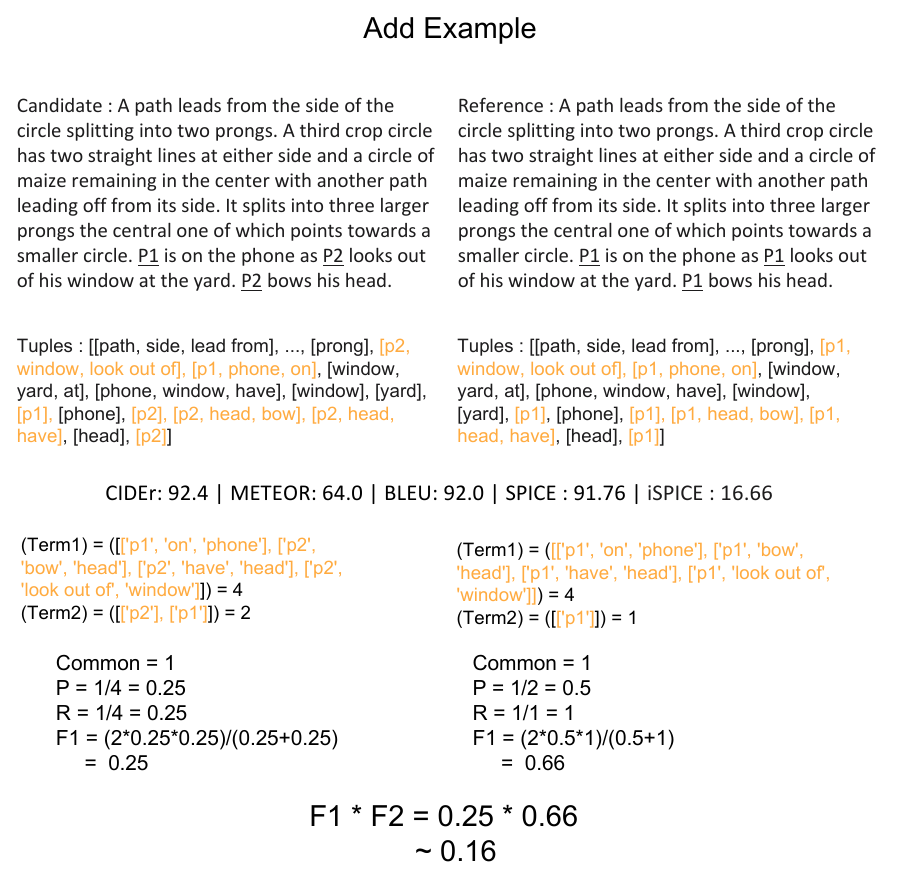}}
\includegraphics[width=0.48\linewidth]{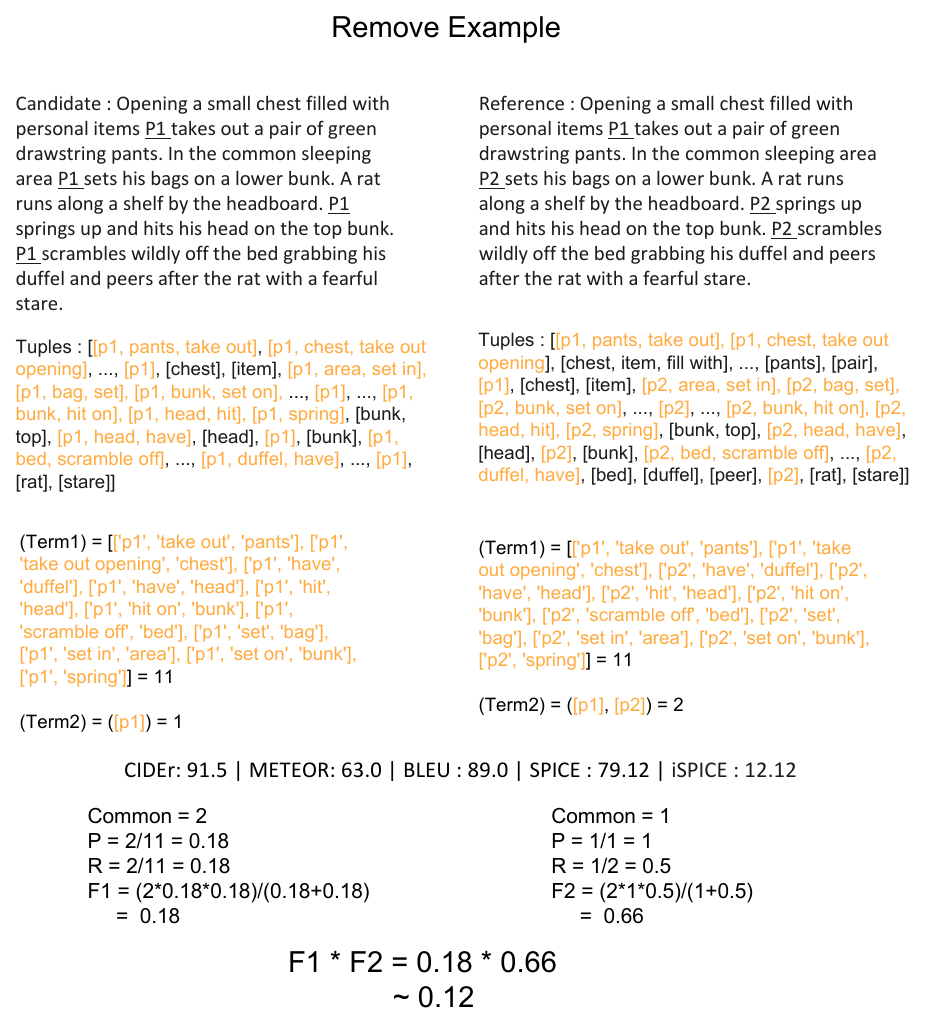} \\
\includegraphics[width=0.48\linewidth]{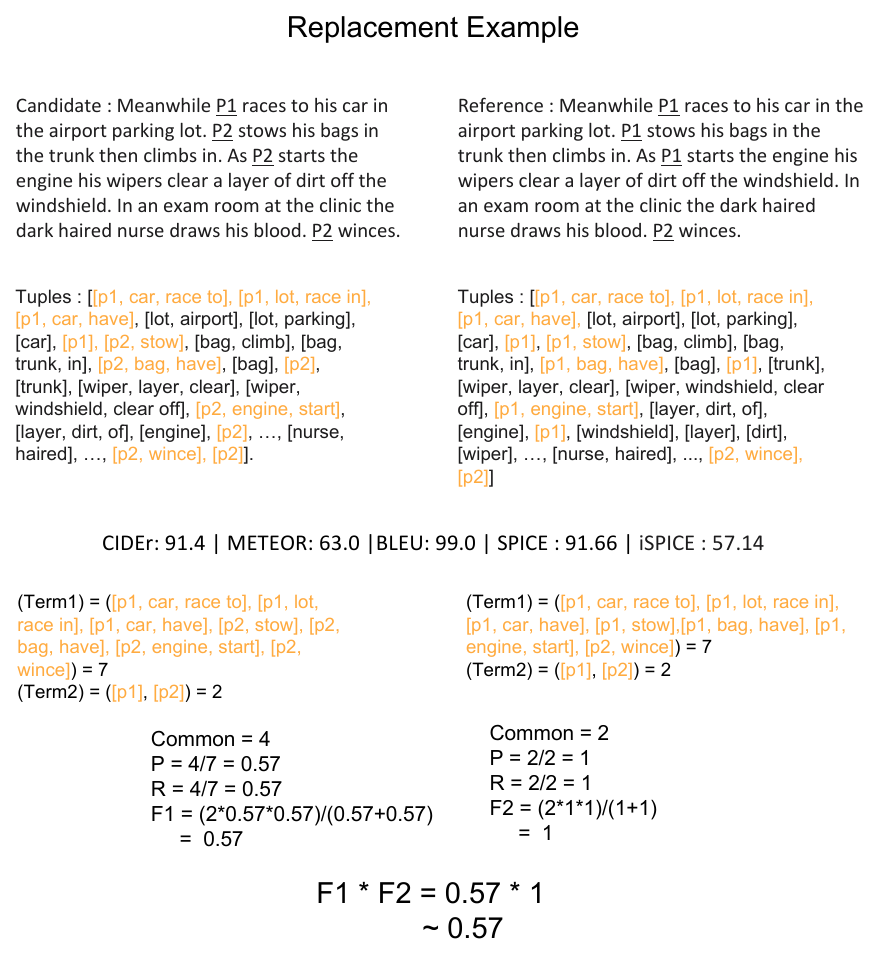}
\caption{We show the effect of identity on captioning metrics using add, remove, and replacement examples.
This corresponds to the validation experiment conducted in Tab.~1 of the main paper.
For each example, the identity labels are underlined in the candidate and reference captionsets.
We also show how \ispice{} works by illustrating the tuples, highlighting tuples with identities, and showing the computation of term 1 (left) and term 2 (right) corresponding to tuples with size $\geq 1$ and $= 1$ respectively.
\ispice{} takes into the account the identity whereas the other metrics show a high score due to high number of n-gram matches.}
\label{fig:supp_ispice}
\end{figure*}

\begin{figure*}[p]
\centering
\includegraphics[width=\linewidth]{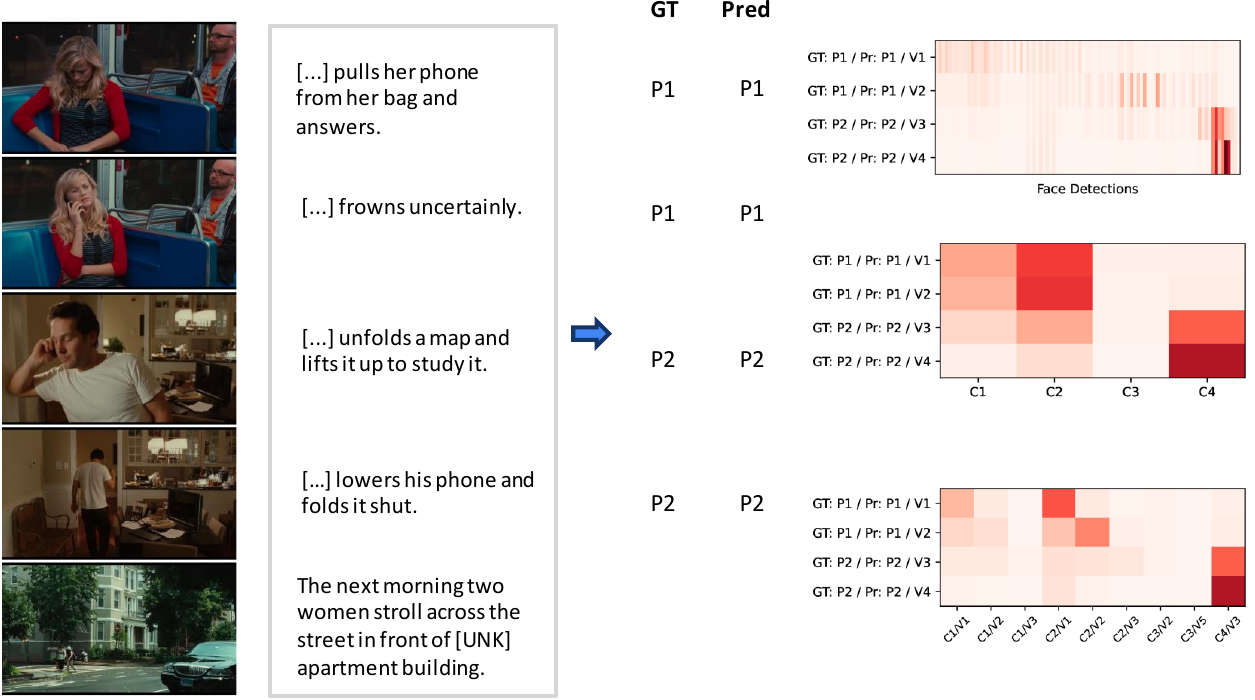}
\includegraphics[width=\linewidth]{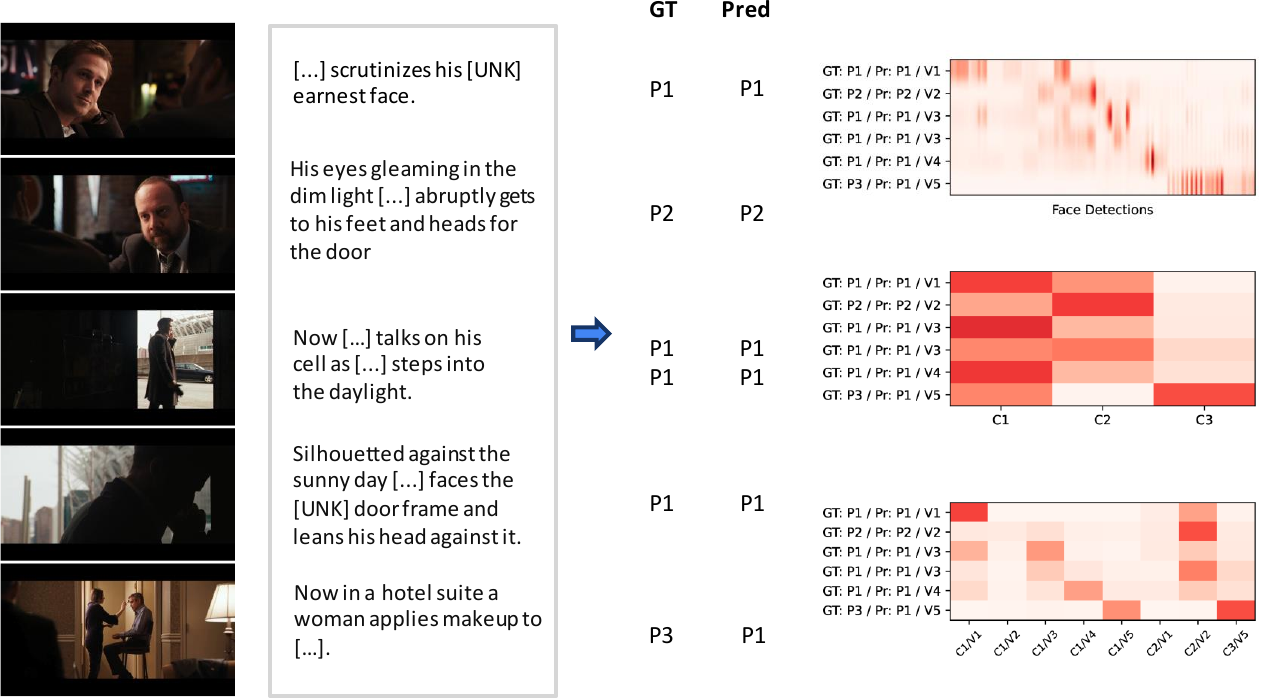}
\caption{Examples from the Fill-in-the-blanks (FITB) task.
On the left, we show one frame from each video of the videoset and the corresponding caption with blanks.
In the middle, we show the ground-truth and predicted person id labels.
On the right, we show the cross-attention maps (face detections, clusters, and clusters by video ids), presented in \cref{fig:supp_attn_fitb}.
We pick the examples corresponding to captionset 3 and 4 of \cref{fig:supp_attn_fitb} for better understanding.
In general, we observe that person predictions depend strongly on the cluster features and their attention.
In some cases, the identity may be difficult to predict as seen in the last row of the second example, where our model predicts P1 instead of P3, even though the attention masks are correctly focusing on C3/V5.}
\label{fig:supp_fitb}
\end{figure*}

\begin{figure*}[p]
\centering
\raisebox{3mm}{\includegraphics[width=0.48\linewidth]{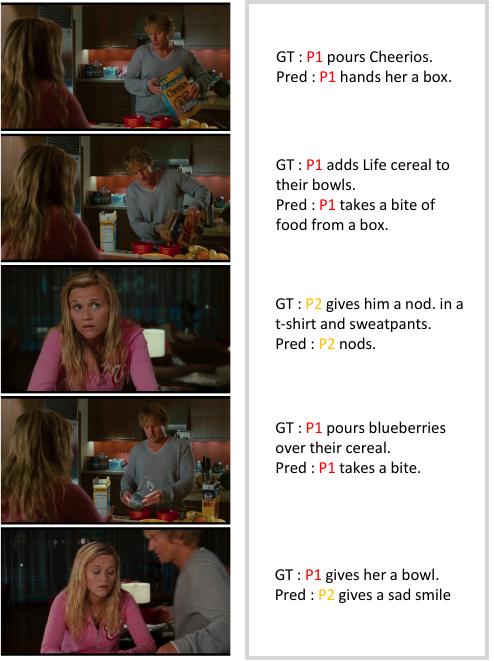}}
\includegraphics[width=0.48\linewidth]{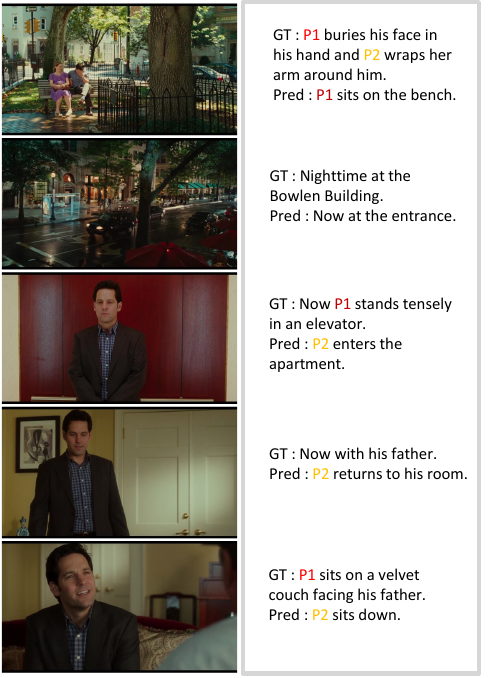}
\caption{The above examples showcase \modelname's ability to perform id-aware captioning.
We see that the predicted captions are quite good, although terse.
While the GT captions tend to be more descriptive in nature, we believe that such behavior may be introduced in the future by incorporating Large Language Models for captioning.}
\label{fig:supp_idawarecap1}
\end{figure*}

\begin{figure*}[p]
\centering
\includegraphics[width=0.48\linewidth]{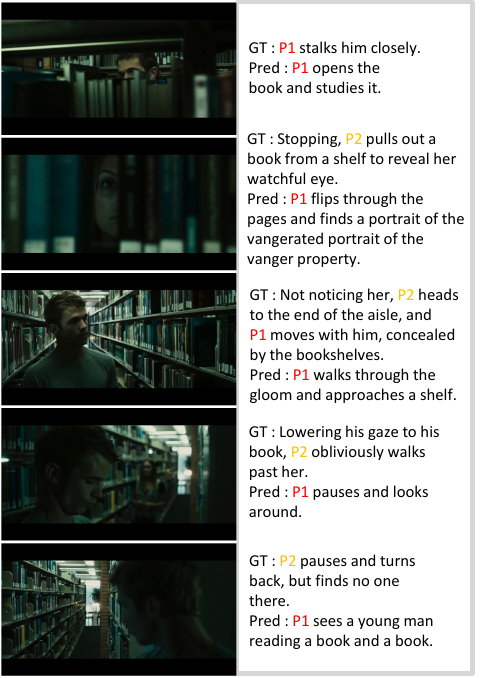}
\includegraphics[width=0.48\linewidth]{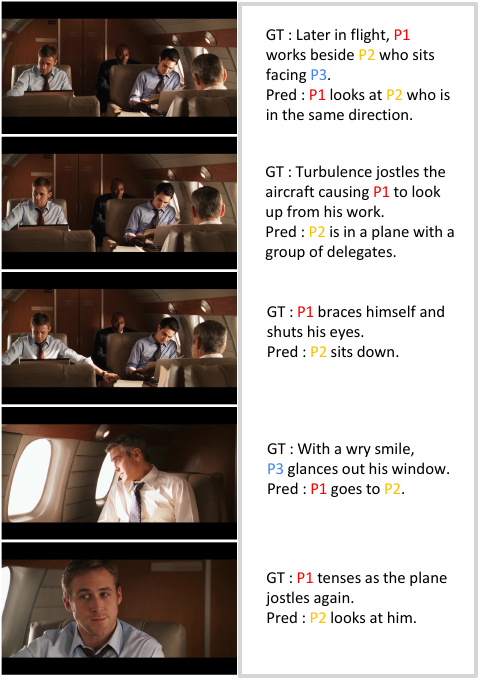}
\caption{The above examples are relatively difficult cases where there are multiple characters involved with lot of drama or action happening in quick succession.
The characters faces are also occluded or partly visible (left example) making it harder to predict identity.
We observe that the predicted captions do not capture the tension (\eg~plane turbulence) and the identities.}
\label{fig:supp_idawarecap2}
\end{figure*}

\section{Qualitative Results}
\label{sec:supp_qualitative}

\paragraph{\ispice{} validation examples.}
To validate our new metric, we propose an experiment that measures similarity between captions when identity names are added, removed, or replaced (Sec.~4 of the main paper).
While the quantitative results favor \ispice, as seen in Tab.~1 of the main paper, we illustrate with examples the process of metric computation in \cref{fig:supp_ispice}.
We observe that the small difference in identity names is captured correctly by \ispice, due to the focus on tuples containing identities, while other metrics do not show this sensitivity.

\paragraph{FITB examples.}
While \cref{fig:supp_attn_fitb} clearly shows the importance of cross-attention scores of detected faces and computed clusters, the challenging visual scenarios are not evident.
We pick two examples (captionset 3 and 4) from \cref{fig:supp_attn_fitb} and pair them together with one frame from each video of the videosets.
\cref{fig:supp_fitb} shows the challenging nature of the videos where characters are often not looking at the camera (example 1 video 1, 3),
the scene is dark,
or the face may not even be visible (example 1 video 4 or  example 2 video 3).
\modelname{} leverages the ability to look at faces and clusters across videos to improve results on the FITB task.

\paragraph{Id-aware captioning examples.}
\cref{fig:supp_idawarecap1} shows 2 examples where our model does relatively well, while \cref{fig:supp_idawarecap2} shows 2 difficult examples where our model makes mistakes.

In the left column of \cref{fig:supp_idawarecap1} we see that the model rightly identifies P1 as the male character and P2 as the female.
The last caption is quite interesting -- while the GT points to P1 giving P2 a bowl, our model predicts that P2 gives a sad smile, which is not wrong.
This also illustrates some of the challenges of evaluating captioning.
In the right column of \cref{fig:supp_idawarecap1}, the predicted caption uses P2 to refer to the man, and is consistent across videos 3, 4, and 5 in the videoset.

In the complex visual example of \cref{fig:supp_idawarecap2} (left), our model assigns P1 to all blanks.
Similarly, in the multi-character example of \cref{fig:supp_idawarecap2} (right), we observe some confusion between characters.
Nevertheless, P2, identified as the man on the left in video 3, is correctly identified for the first 3 videos.
The model is also able to predict that they are on a plane (caption for video 2).
Nevertheless, these examples illustrate the challenges of id-aware captioning.
As future work, they also highlight the need to evaluate visual grounding of the identities beyond captioning performance.

\section{Limitation and Future Work}
\label{sec:supp_limitations}

One limitation of our work, inherited from the task definition in LSMDC, is restricting videosets to local groups of 5 videos.
In the future, we would like to extend this to larger videosets, perhaps spanning the entire movie.
However, the approach will need to be modified to operate on full movies as:
(i)~providing features of all movie frames as decoder memory creates a huge number of embeddings;
(ii)~face clustering across the entire movie could be error-prone; and
(iii)~auto-regressively generating one caption at a time for hundreds of clips seems challenging, as the model needs to be cognizant of all previously generated captions.
We believe that a hierarchical model that builds from shots to scenes to the full movie may be more appropriate here.

Second, the tasks for FITB and full captioning do not learn at the same pace, and choosing a single best checkpoint for both may be difficult.
We posit that the user may choose two checkpoints, one for each task.
Furthermore, we observe that by weighing the FITB and full captioning losses appropriately, additional performance improvements can be achieved for one task at the cost of the other task.

We have also not considered using external knowledge or pre-trained large language models (LLMs) or vision-language models (VLMs) built for captioning.
We believe that it is interesting to learn what can be achieved by training on LSMDC alone.
As seen in multiple examples throughout \cref{sec:supp_qualitative}, \modelname{} does perform quite well given the challenging scenarios.

\end{document}